\documentclass[a4paper,fleqn]{cas-dc}

\usepackage[numbers]{natbib}

\def\tsc#1{\csdef{#1}{\textsc{\lowercase{#1}}\xspace}}
\tsc{WGM}
\tsc{QE}

\usepackage[nolist]{acronym}

\usepackage{tabularx,booktabs}
\newcolumntype{Y}{>{\centering\arraybackslash}X}
\newcolumntype{Z}{>{\arraybackslash}X}
\usepackage{booktabs}  
\usepackage[binary-units=true]{siunitx}
\usepackage{textcomp}
\usepackage{bm}  
\usepackage{interval}

\usepackage{makecell}
\usepackage{lipsum}

\usepackage{tikz}
\usepackage{pgfplots}
\pgfplotsset{compat=1.16}
\usetikzlibrary{backgrounds,arrows,automata,shapes,positioning,calc,through}
\pgfdeclarelayer{background}
\pgfdeclarelayer{foreground}
\pgfsetlayers{background,main,foreground}

\usetikzlibrary{fit}

\tikzset{
  highlight/.style={
    circle,
    draw=blue, very thick,
    minimum height=2.5em,
    text centered,
    inner sep = 0pt,
  },
  highlight2/.style={
    circle,
    draw=red, very thick,
    minimum height=5.8em,
    text centered,
    inner sep = 0pt,
  },
  state/.style={
    rectangle,
    draw=black, very thick,
    minimum height=1.0em,
    text centered,
  },
  final_state/.style={
    rectangle,
    rounded corners,
    draw=black, very thick,
    minimum height=2em,
    text centered,
    dashed
  },
  initial_state/.style={
    rectangle,
    double=white,
    double distance=1pt,
    inner sep=2pt,
    draw=black, very thick,
    minimum height=2em,
    text centered,
  },
  point/.style={
    circle,
    inner sep=0pt,
    minimum size=3pt,
    fill=red
  }
}

\usepackage{pstricks}
\usepackage{tikz-3dplot}

\renewcommand\vec{\bm}
\newcommand\mat{\mathbf}

\newcommand{\framess}[3]{{#3}^{#1}_{#2}}
\usepackage{cancel}

\usepackage{mathtools}

\newcommand{\DOI}{} 

\usepackage[placement=top,vshift=-7]{background}
\SetBgScale{1.0}
\SetBgContents{Preprint version. This work has been submitted to Elsevier for possible publication.}
\SetBgColor{black}
\SetBgAngle{0}
\SetBgOpacity{1.0}

\begin{document}
\emergencystretch 3em

\let\WriteBookmarks\relax
\def\floatpagepagefraction{1}
\def\textpagefraction{.001}

\shorttitle{Degradation-Aware Cooperative Multi-Modal GNSS-Denied Localization}    

\shortauthors{Pritzl et al.}  

\title [mode = title]{Degradation-Aware Cooperative Multi-Modal GNSS-Denied Localization Leveraging LiDAR-Based Robot Detections}  

\tnotemark[1] 

\tnotetext[1]{This work was supported by CTU grant no. SGS26/077/OHK3/1T/13, by the Czech Science Foundation (GAČR) under research project No. GA26-22606S, and by the European Union under the project Robotics and advanced industrial production (reg. no. CZ.02.01.01/00/22\_008/0004590).} 

%

\author[1]{Václav Pritzl}[orcid=0000-0002-7248-6666]
\cormark[1]
\ead{vaclav.pritzl@fel.cvut.cz}
\credit{conceptualization, methodology, software, validation, formal analysis, investigation, data curation, writing - original draft, writing - review \& editing, visualization}
\affiliation[1]{organization={Multi-robot Systems Group, Department of Cybernetics, Faculty of Electrical Engineering, Czech Technical University in Prague},
            addressline={Technická 2}, 
            city={Prague},
            postcode={166 27}, 
            country={Czech Republic}}

\author[2]{Xianjia Yu}[orcid=0000-0002-9042-3730]


\ead{xianjia.yu@utu.fi}


\credit{resources, investigation, writing - review \& editing}

\affiliation[2]{organization={Turku Intelligent Embedded and Robotic Systems (TIERS) Lab, University of Turku},
            city={Turku},
            postcode={20520}, 
            country={Finland}}

\author[2]{Tomi Westerlund}[orcid=0000-0002-1793-2694]


\ead{tovewe@utu.fi}


\credit{resources, investigation, writing - review \& editing}


\author[1]{Petr Štěpán}[orcid=0000-0002-7444-3264]


\ead{petr.stepan@fel.cvut.cz}


\credit{supervision, writing - review \& editing}

\author[1]{Martin Saska}[orcid=0000-0001-7106-3816]


\ead{martin.saska@fel.cvut.cz}


\credit{funding acquisition, project administration, supervision, writing - review \& editing}

\cortext[1]{Corresponding author}



\begin{abstract}
  Accurate long-term localization using onboard sensors is crucial for robots operating in \ac{GNSS}-denied environments.
While complementary sensors mitigate individual degradations, carrying all the available sensor types on a single robot significantly increases the size, weight, and power demands.
  Distributing sensors across multiple robots enhances the deployability but introduces challenges in fusing asynchronous, multi-modal data from independently moving platforms.
  We propose a novel adaptive multi-modal multi-robot cooperative localization approach using a factor-graph formulation to fuse asynchronous \ac{VIO}, \ac{LIO}, and 3D inter-robot detections from distinct robots in a loosely-coupled fashion.
The approach adapts to changing conditions, leveraging reliable data to assist robots affected by sensory degradations.
  A novel interpolation-based factor enables fusion of the unsynchronized measurements.
  \ac{LIO} degradations are evaluated based on the approximate scan-matching Hessian.
  A novel approach of weighting odometry data proportionally to the Wasserstein distance between the consecutive \ac{VIO} outputs is proposed.
  A theoretical analysis is provided, investigating the cooperative localization problem under various conditions, mainly in the presence of sensory degradations.
  The proposed method has been extensively evaluated on real-world data gathered with heterogeneous teams of an \ac{UGV} and \acp{UAV}, showing that the approach provides significant improvements in localization accuracy in the presence of various sensory degradations.
\end{abstract}




\begin{keywords}
  Multi-Robot Systems \sep Cooperative Localization \sep Multimodal Localization \sep Factor Graph \sep Unmanned Aerial Vehicle \sep Unmanned Ground Vehicle \sep LiDAR
\end{keywords}

\maketitle


\section*{Supplementary Material}
\noindent\textbf{Video:} \url{https://mrs.fel.cvut.cz/coop-mm-localization}


\section{Introduction}

\label{sec:introduction}

\acresetall

In \ac{GNSS}-denied environments, fusing different localization modalities is crucial to provide robustness to various environmental challenges~\cite{ebadiPresentFutureSLAM2024}.
Visual-based localization requires cheap and light-weight sensors, but it is sensitive to illumination changes and texture-less environments.
\ac{lidar}-based localization exhibits better accuracy and works in challenging lighting conditions, but requires heavy power-consuming sensors, significant geometric structure in the environment, and presence of objects in the relatively small range of 3D \acp{lidar}.

Multi-robot localization methods enable the distribution of different sensors on different robots in different parts of the environment.
In such a case, all robots do not need to carry all the sensors to be able to fully function in every specific scenario.
A \ac{lidar}-carrying robot can provide accurate localization in an area with sufficient geometric structure, while a camera-equipped robot can take care of the localization in a place without geometric features but with enough visual texture.
Moreover, multi-robot localization enables the robot team to cooperate to fulfill the desired task, such as cooperative mapping, sensing, inspection, or search and rescue.

\begin{figure}[t]
  \centering
      \includegraphics[width=1.0\linewidth, trim=0cm 0cm 0cm 0cm, clip=true]{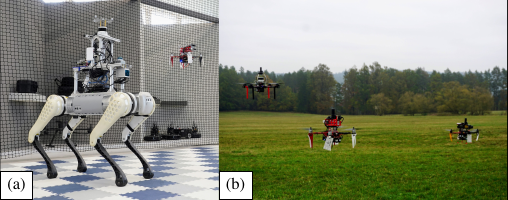}

  \caption{(a) A \ac{lidar}-equipped \ac{UGV} with a single camera-carrying \ac{UAV} and (b) one \ac{lidar}-equipped \ac{UAV} cooperating with two camera-equipped \acp{UAV}.}
  \label{fig:motivation}
\end{figure}

Multi-modal multi-robot localization is crucial for enabling truly heterogeneous robot teams to operate in various environmental conditions (see Fig.~\ref{fig:motivation}), but it comes with a set of its own unique challenges.
In a multi-modal multi-robot localization scenario, we need to fuse data from sensors carried by multiple different robots, which are independently moving with respect to each other and communicating over a network with possibly limited bandwidth.
The data are asynchronous, and the sensory degradations can cause certain parts of the estimated state to be unobservable.

Multi-modal localization methods can be loosely-coupled, processing the modalities separately into distinct estimates and fusing them afterwards, or tightly-coupled, jointly processing the modalities at the feature / sensory data level.
While tightly-coupled methods are generally more accurate, applying such methods to multi-robot scenarios is challenging due to the necessity to transmit more data between the robots and due to the data being obtained from mutually distant and moving locations.

In contrast, a loosely-coupled fusion scheme requires comparatively lower communication bandwidth and enables modularity, providing the option to easily swap between the specific algorithms processing each sensory modality.
However, such a loosely-coupled fusion method needs to be able to efficiently evaluate the reliability of each input to adaptively fuse them to provide an accurate location estimate at all times.


\subsection{Problem Statement}
\label{sec:problem_statement}

\begin{table*}[htbp]
\small
\begin{center}
  \caption{Mathematical notation and nomenclature.}
    \begin{tabularx}{0.95\linewidth}{p{2.5cm} X} 
    \toprule 
    \textbf{Symbol} & \textbf{Meaning} \\ \midrule
    $a$ & scalar value $a$ \\
      $\vec{a}$ & vector $\vec{a}$ \\
      $\mat{A}$ & matrix $\mat{A}$ \\
      $\mathcal{A}$ & set $\mathcal{A}$ \\
    $A$ & reference frame / robot $A$ \\
    $SE(3)$ & special Euclidean group \\
    $SO(3)$ & special orthogonal group \\
    $\framess{A}{B}{\mat{T}}$ & $SE(3)$ pose of $B$ in frame $A$, matrix transforming vectors from $B$ to $A$ \\
    $\framess{A}{B}{\hat{\mat{T}}}$ & estimated $SE(3)$ pose of $B$ in frame $A$ \\
    $\tilde{\mat{T}}$ & pose obtained by applying tangent-space perturbation to pose $\mat{T}$ \\
    $\framess{A}{B}{\mat{R}}$ & $SO(3)$ rotation of $B$ in frame $A$ \\
    $\mat{R}_z(\theta)$ & rotation matrix about the $z$-axis by angle $\theta$  \\
    $\framess{W}{A}{\mat{R}}(\alpha,\beta)$ & constraint on the roll $\alpha$ and pitch $\beta$ of $A$ in frame $W$ \\
    $\framess{A}{B}{\vec{t}}$ & translation of $B$ in frame $A$ \\
    $\framess{A}{B}{\vec{d}}$ & detection of $B$ in frame $A$ \\
    $\vec{\xi}_A$ & perturbation vector on the tangent space associated with the pose $\framess{W}{A}{\mat{T}}$; applied on the right side of $\framess{W}{A}{\mat{T}}$; orientation is expressed before translation \\
      $\mathrm{Exp}:\nolinebreak\mathbb{R}^6\nolinebreak\to\nolinebreak SE(3)$ & exponential map, mapping from the local vector space to the Lie group $SE(3)$ \\
      $\mathrm{Log}:\nolinebreak SE(3)\nolinebreak\to\nolinebreak\mathbb{R}^6$ & logarithm map, mapping from the Lie group $SE(3)$ to local vector space \\
    $\mat{\Sigma}$ & covariance matrix \\
    $\mathcal{N}(\vec{\mu},\mat{\Sigma})$ & multivariate normal distribution with mean $\vec{\mu}$ and covariance matrix $\mat{\Sigma}$ \\
    $W_2(\mat{\Sigma}_1, \mat{\Sigma}_2)$ & 2-Wasserstein distance between $\mat{\Sigma}_1$ and $\mat{\Sigma}_2$ \\
    $\mathrm{Tr}\mat{A}$ & trace of matrix $\mat{A}$ \\
    $\left[ \framess{A}{B}{\mat{T}} \right]_\mathrm{tr}$ & translational component of pose $\framess{A}{B}{\mat{T}}$ \\
    $\mat{J}$ & Jacobian matrix \\
    $[\vec{x}]_\times$ & skew-symmetric matrix constructed from $\vec{x}$ \\
    $\mat{J}_{r}\big|_{\vec{x}}$ & right Jacobian of $SE(3)$ evaluated at $\vec{x}$  \\
    $\mat{Ad_T}$ & adjoint of $\mat{T}$  \\
    $\mat{I}_6$ & identity matrix of size $6\times6$\\
    $\left[\vec{a}\right]_{x}$ & $x$-component of the vector $\vec{a}$ \\
    $\vec{a}^\bot$ & vector orthogonal to vector $\vec{a}$\\
    \bottomrule
\end{tabularx}
  \label{tab:notation}
\end{center}
\end{table*}

\begin{figure}[t]
\centering
  \includegraphics[width=0.9\linewidth, trim=0.0cm 0.0cm 0.0cm 0.0cm, clip=true]{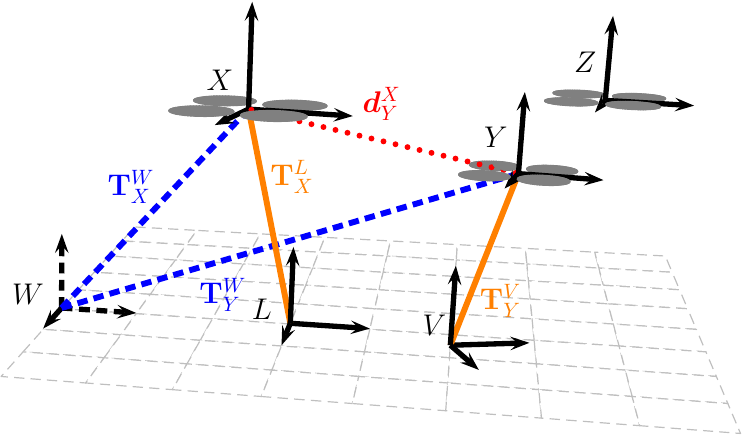}
  \caption{Robot $X$ is localized in local frame $L$ using a \ac{LIO} algorithm. Robot $Y$ is localized in a local frame $V$ using a \ac{VIO} algorithm.
  Robot $X$ detects the relative 3D position of robot $Y$.
  Robot $Z$ represents another \ac{VIO}-utilizing robot detected by robot $X$.
  $W$ denotes the world reference frame of the cooperative localization algorithm.
  All the reference frames are gravity-aligned.
  Blue dashed lines represent the estimated variables $\framess{W}{X}{\mat{T}}, \framess{W}{Y}{\mat{T}}$.
  Orange solid lines represent poses $\framess{L}{X}{\mat{T}},\framess{V}{Y}{\mat{T}}$ provided by the individual odometry algorithms.
  The red dotted line represents the relative detection $\framess{X}{Y}{\vec{d}}$ of robot $Y$ by robot $X$.}
\label{fig:drones_tikz}
\end{figure}

\noindent We tackle the problem of adaptive loosely-coupled multi-modal multi-robot localization.
The problem is illustrated in Fig.~\ref{fig:drones_tikz}.
Table~\ref{tab:notation} describes the mathematical notation and nomenclature used throughout the paper.
We focus on the case of cooperative localization in a team of one \ac{lidar}-carrying robot and one or more camera-equipped robots.
  A \ac{lidar}-carrying robot $X$ is localized in local frame $L$ using a \ac{LIO} algorithm, providing pose $\framess{L}{X}{\mat{T}}$.
A camera-equipped robot $Y$ is localized in local frame $V$ using a \ac{VIO} algorithm, providing pose $\framess{V}{Y}{\mat{T}}$.
Robot $X$ detects the 3D position $\framess{X}{Y}{\vec{d}}$ of robot $Y$ w.r.t. the gravity-aligned body frame of robot $X$.
  We assume that all the frames of reference are aligned to the gravity vector as all robots are equipped with \acp{IMU}, enabling the estimation of the gravity vector direction and the 6-\ac{DOF} poses in the gravity-aligned reference frames.

The task tackled in this paper is to utilize the \ac{LIO} / \ac{VIO} data from separate robots and inter-robot detections to periodically estimate the poses $\framess{W}{X}{\mat{T}}, \framess{W}{Y}{\mat{T}}$ of each robot in the common reference frame $W$, while minimizing localization error with respect to the true robot poses.
  As all the reference frames are gravity-aligned, all the estimated poses have 4 \acp{DOF} with the roll and pitch angles set to zero.
  As the odometry algorithms of the individual robots are capable of producing full 6-\ac{DOF} pose estimates in gravity-aligned reference frames, the full 6-\ac{DOF} poses in a common reference frame can be obtained by simply transforming the output of the cooperative localization algorithm using the tilt obtained from each respective odometry output.
  Similarly, the inter-robot detections in a gravity-aligned reference frame are obtained by transforming the detections in the tilted body frame using the current tilt of the robot obtained from the odometry output.

The system times of all robots are assumed to be synchronized throughout the estimation process.
The odometry data and detections are produced asynchronously.
The inter-robot detections are anonymous, requiring the approach to solve the association problem.
No global measurements, i.e., no \ac{GNSS} data, are available to the algorithm.
The robots are capable of explicit communication.

The \ac{LIO} algorithm is based on solving the scan matching problem, optimizing a point-to-plane / point-to-line cost function, enabling the option to analyze the approximate Hessian to evaluate scan matching degeneracy.
The \ac{VIO} is based on a Kalman filter, outputting uncertainty estimates in the form of covariance matrices.

We focus on the specific case of \ac{LIO} and \ac{VIO}-utilizing robots to highlight the significant applicability of the approach to heterogeneous robot teams, but it is worth mentioning that the proposed method is applicable to other odometry sources as well, assuming that they provide similar capabilities of evaluating their uncertainty.
Similarly, though we focus on \ac{lidar}-based detections between the robots, the method is applicable to other relative localization methods capable of producing accurate 3D detections of neighboring robots.

\subsection{Related Work}\label{sec:relatedwork}

\subsubsection{Multi-Modal Localization on a Single Robot}
Most single-robot fusion approaches are not easily applicable to a multi-robot scenario.
Many approaches operate in a coarse-to-fine manner, obtaining the estimated pose of the robot body with one modality and using the result as a prior to refine the estimate with another modality~\cite{zhangLaserVisualInertial2018b, khattakComplementaryMultiModal2020}.
In~\cite{zhaoSuperOdometryIMUcentric2021a}, the authors proposed an \ac{IMU}-centric multi-modal odometry based on a factor graph.
In their approach, they constrained the \ac{IMU} prediction using relative pose factors and detected possible sensory degenerations by analyzing the information matrices of the respective optimization problems for each modality.
Many single-robot multi-modal approaches use a tightly-coupled scheme~\cite{graeterLIMOLidarMonocularVisual2018, shanLVISAMTightlycoupledLidarVisualInertial2021a, linR33LIVERobustRealTime2024, leeMINSEfficientRobust2023, yuanSRLIVOLiDARInertialVisualOdometry2024, zhangLVIOFusionTightlyCoupledLiDARVisualInertial2024, zhengFASTLIVO2FastDirect2025}, fusing \ac{lidar}, visual, and inertial measurements in a single estimator without dividing the process into completely separate modality-specific subsystems.
In~\cite{xuIntermittentVIOAssistedLiDAR2025}, the authors utilize intermittent \ac{VIO} data when \ac{lidar} degeneracy is detected, while aiding \ac{VIO} by using \ac{lidar} points as additional features.
In~\cite{wangMCLIVOLowdriftLiDARinertialvisual2025a}, a factor-graph-based \ac{lidar}-visual-inertial odometry was proposed, fusing data from different modalities as relative poses and associating depth information from the \ac{lidar} data to the visual information.
Despite the advantages of these methods in a single robot scenario, a coarse-to-fine approach, tight-coupling of sensor data, associating depth from \ac{lidar} data with visual features, or centering the approach around a single \ac{IMU} odometry are not easily applicable to a multi-robot scenario, where each sensory modality is placed on a separate robot, the robots need to wirelessly communicate their data, and the changing transformation between the robots is obtained through noisy detections.
Therefore, we focus on a parallel loosely-coupled architecture, running completely separate odometry algorithms on each robot, converting them to relative poses, and weighting the individual poses based on their estimated reliability.

\subsubsection{Evaluating the Uncertainty of a Sensory Modality}
To achieve accurate and robust performance, loosely-coupled fusion methods must be able to detect sensory degradations and adapt the fusion process accordingly.
For \ac{lidar}-fusing methods based on minimizing the point-to-plane cost function, the standard solution is to analyze the eigenvalues of the approximate Hessian of the optimization problem and perform thresholding or calculate a covariance matrix from the approximate Hessian~\cite{zhangDegeneracyOptimizationbasedState2016a, zhangLaserVisualInertial2018b, khattakComplementaryMultiModal2020, tunaInformedConstrainedAligned2024a, tunaXICPLocalizabilityAwareLiDAR2024b}.
In our method, we obtain a binary evaluation of the reliability of \ac{lidar}-based localization by thresholding the minimum eigenvalue of the approximate Hessian matrix.

For visual-based methods, the usual approaches for evaluating uncertainty and detecting sensory degradation include analyzing the number of visual features~\cite{wangMCLIVOLowdriftLiDARinertialvisual2025a, horynaFastSwarmingUAVs2024}, thresholding the estimated \ac{IMU} biases~\cite{wangMCLIVOLowdriftLiDARinertialvisual2025a}, analyzing the determinant of the covariance matrix~\cite{khattakComplementaryMultiModal2020} in Kalman filter-based algorithms, thresholding on the maximum change of odometry output with respect to \ac{IMU} odometry~\cite{wangMCLIVOLowdriftLiDARinertialvisual2025a}, or analyzing the information matrix of the optimization problem~\cite{zhaoSuperOdometryIMUcentric2021a}.
However, even a small amount of visual features may provide sufficient constraints, and thus, the number of features may not sufficiently correlate with the localization error.
Thresholding \ac{IMU} biases may provide enough information about the complete loss of localization, but not about finer uncertainty changes.
In a Kalman filter-based \ac{VIO}, analyzing the covariance matrices is the obvious choice.
We need to obtain the uncertainty of the relative poses between consecutive filter outputs.
However, the cross-covariances between the consecutive filter outputs are unknown.
Ignoring cross-covariances when calculating relative pose covariance would lead to degradation of the uncertainty~\cite{mangelsonCharacterizingUncertaintyJointly2020}.

In this work, we propose a novel approach of weighting relative poses calculated from the outputs of a Kalman filter-based algorithm proportionally to the Wasserstein distance between the probability distributions of consecutive filter outputs.
Such an approach is compatible with any Kalman filter-based method without requiring any modifications to the odometry method itself.

\subsubsection{Multi-Robot Cooperative Localization}

Multi-robot \ac{SLAM} algorithms usually rely on inter- and intra-robot loop closures, obtaining relative transformations between the different robot frames by aligning mutually observed features, viewpoints, or partial maps.
The architecture usually consists of a front-end odometry combined with a back-end pose graph optimization running on a server or running on each robot in a decentralized fashion.
Most of such approaches focus on a single exteroceptive modality.
In~\cite{zhongDCLSLAMDistributedCollaborative2024, huangDiSCoSLAMDistributedScan2022a, changLAMP20Robust2022}, such methods are proposed for 3D \ac{lidar} sensors.
In~\cite{zhongCoLRIOLiDARRangingInertialCentralized2024}, pairwise \ac{UWB} measurements are fused with the \ac{lidar} odometry and \ac{lidar}-based loop closures.
In~\cite{schmuckCOVINSVisualInertialSLAM2021, tianKimeraMultiRobustDistributed2022a, birdDVMSLAMDecentralizedVisual2025}, visual-based multi-robot \ac{SLAM} methods are proposed, combining \ac{VIO} front-ends with visual-based place recognition or map-based loop closures.
The authors of~\cite{xuOmniSwarmDecentralizedOmnidirectional2022} fused \ac{VIO} ego-motion estimates with \ac{UWB} ranges, visual detections of cooperating \acp{UAV}, and map-based loop closures.
However, all of these methods are limited in the sense that they utilize only a single main sensory modality, either visual or \ac{lidar} data.
Such an approach lacks the adaptiveness that can be provided by a multi-modal method.
In~\cite{heGroundAerialCollaborative2021}, a collaborative localization and mapping method for a ground-aerial team is proposed, but each robot in the team needs to be equipped with both a \ac{lidar} and a camera.
In~\cite{lajoieSwarmSLAMSparseDecentralized2024}, a \ac{SLAM} method for a \ac{UAV} swarm fusing both \ac{lidar} and visual data is proposed.
However, all robots need to carry both \ac{lidar} and camera sensors, as the inter-robot loop closures are only obtained for each modality separately.
Obtaining loop closures from a combination of different modalities is complicated, as the sensors inherently capture different types of information.
Furthermore, place-recognition-based approaches require significant communication bandwidth to perform the loop closures.

Utilizing direct detections of cooperating robots enables straightforward relative localization without requiring the robots to use the same sensory modalities and minimizes the communication requirements.
In~\cite{xuOmniSwarmDecentralizedOmnidirectional2022}, visual detections were fused in a graph-based approach in addition to loop closures, \ac{VIO} odometry, and \ac{UWB} measurements.
The approach relied on broadcasting the \ac{UWB} measurements and odometry data at \SI{100}{Hz}, converting the timestamps of all the measurements to the closest \ac{UWB} timestamps, and using accurate \ac{VIO} odometry data for propagating the detections and map-based measurements to the timestamps of the estimation variables.
In contrast, our approach is built around a novel interpolation-based quaternary factor representing the inter-robot detections.
Such an approach enables us to have the estimation variables at arbitrary times, independent of the times when the detections were produced, and removes the need for high-rate odometry data for propagating the measurements, reducing the communication and computational requirements.
In our experiments, the odometry data were utilized at the rate of \SI{2}{Hz}.

In~\cite{horynaFastSwarmingUAVs2024}, ultraviolet markers were utilized for camera-based detection of cooperating \acp{UAV} in a multi-robot state estimation method.
In \cite{zhuSwarmLIO2DecentralizedEfficient2025}, the authors proposed a tightly-coupled decentralized \ac{lidar}-based odometry for a swarm of \acp{UAV} utilizing \ac{lidar}-based detections of reflective markers on board the \acp{UAV}.
However, the aforementioned works all relied primarily on a single sensory modality, either \ac{lidar} or camera-based.
In \cite{pritzlFusionVisualInertialOdometry2023}, \ac{lidar}-based detections were utilized for relative localization in a heterogeneous team of a \ac{lidar}-equipped and camera-equipped \ac{UAV}.
However, the fusion approach did not deal with variable uncertainties of the localization inputs and always treated the \ac{lidar}-based localization as perfectly reliable.
Therefore, it would fail in the presence of \ac{lidar} degradations.
In \cite{Spasojevic-RSS-23}, the authors proposed an active collaborative localization method for a \ac{UGV}-\ac{UAV} team.
The \acp{UGV} were detected from cameras on board the \acp{UAV} and used as landmarks to improve the localization of the \acp{UAV}.
The work primarily dealt with the optimal placement of the \acp{UGV} in the environment.
However, the \acp{UGV} were only utilized as static landmarks, and the work did not deal with improving localization performance in the other way around, i.e., improving localization of the \acp{UGV} by the \acp{UAV}.

In our work, we utilize direct 3D detections of the cooperating robots, obtained from \ac{lidar} data.
In contrast to the state-of-the-art methods, we focus on a scenario where the multi-robot team combines multiple different sensory modalities, but each robot is equipped with only one distinct exteroceptive sensor type, e.g., one robot is only equipped with a \ac{lidar}, while another robot is only equipped with a camera.
Our approach adapts to changing sensory degradations in the environment and does not rely on the assumption that a single modality is always reliable or always more accurate than the other one.


\subsection{Contributions}\label{sec:contributions}

\noindent The contributions of this work are summarized as:
\begin{itemize}
  \item A novel adaptive loosely-coupled multi-modal multi-robot fusion method for cooperative localization.
    The method utilizes direct 3D detections of cooperating \acp{UAV}, enabling us to efficiently fuse localization outputs from different sensory modalities without the need to perform place recognition on the sensory data.
    The detections are represented by a novel interpolation-based quaternary factor, enabling efficient fusion of data from unsynchronized sources.
    \item A novel approach of weighting relative pose factors proportionally to the Wasserstein distance of consecutive outputs of a Kalman filter-based algorithm, enabling the method to utilize information about the variable uncertainty of the odometry without degrading the estimate due to unknown cross-covariances.
    \item Theoretical observability analysis of the problem of cooperative localization based on odometry estimates and relative 3D detections with respect to various assumptions and odometry degradations.
\end{itemize}
The proposed method was extensively evaluated on real-world datasets gathered with a \ac{UGV}-\ac{UAV} and \ac{UAV}-only team.
The accuracy of the approach was quantitatively evaluated with respect to motion-capture and \ac{RTK} ground truth.
The evaluation of the method included an ablation study on the use of the Wasserstein-distance-based weighting and challenging scenarios subject to various sensory degradations.



\section{Multi-Robot Localization Method}
\label{sec:system_model}

\noindent We formulate the cooperative localization problem as a factor graph~\cite{dellaertFactorGraphsRobot2017} with the estimated robot poses, priors, and measurements represented by multivariate Gaussian distributions.
As the distinct measurement sources are unsynchronized, we utilize a novel interpolation-based factor representing the inter-robot detections.
The odometry measurements are adaptively weighted based on the reliability of each robot's odometry.

\subsection{Factor Graph-Based Formulation}

\begin{figure}[t]
  \centering
  \input{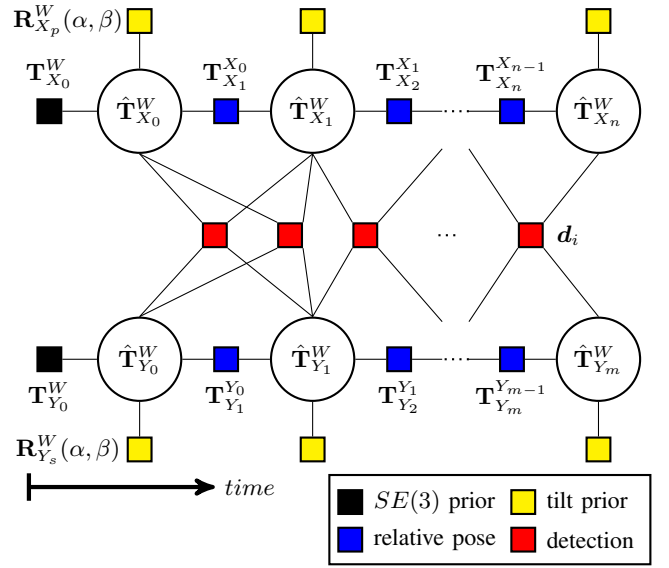}
  \caption{Factor graph representation of the cooperative localization problem for two robots.
  Circles represent the estimated variables.
  Squares represent the factors, i.e., the measurements and prior information.
  Relative poses of robot $X$ are obtained from a \ac{LIO} algorithm, while relative poses of robot $Y$ are provided by a \ac{VIO} algorithm.%
  }

  \label{fig:factor_graph}
\end{figure}

\noindent The factor graph for the cooperative localization problem for two robots is illustrated in Fig.~\ref{fig:factor_graph}.
The problem is modeled by a set of estimated $SE(3)$ variables $\hat{\mathcal{X}} = \{\framess{W}{X_0}{\hat{\mat{T}}}, \framess{W}{X_1}{\hat{\mat{T}}}, \dots, \framess{W}{X_n}{\hat{\mat{T}}}, \framess{W}{Y_0}{\hat{\mat{T}}}, \framess{W}{Y_1}{\hat{\mat{T}}}, \dots, \framess{W}{Y_m}{\hat{\mat{T}}}\}$ representing the poses of the robots at specific moments in time and a set of factors representing the measurements.
The factors include prior factors, the $SE(3)$ relative pose factors representing the relative movement of each robot obtained from an odometry algorithm, and the $\mathbb{R}^3$ inter-robot position detection factors.
Additional robots detected by robot $X$ would be incorporated analogously, by connecting the corresponding set of estimated robot poses to the poses of robot $X$ through the 3D position detection factors.
Let $\mathcal{L} = \{ \framess{X_{k}}{X_{k+1}}{\mat{T}} \mid k=0,\dots n-1 \}$ be the set of $n$ \ac{LIO} relative poses of robot $X$, $\mathcal{V} = \{ \framess{X_{l}}{X_{l+1}}{\mat{T}} \mid l=0,\dots m-1 \}$ be the set of $m$ \ac{VIO} relative poses of robot $Y$, and $\mathcal{D} = \{ \vec{d}_i \mid i = 0,\dots p \}$ be the set of $p$ inter-robot detections.
Let $\mathcal{X}_i, \mathcal{X}_j, \mathcal{X}_k, \mathcal{X}_l, \mathcal{X}_q,$ be the subsets of variables connected to the $i$-th inter-robot detection factor, $j$-th $SE(3)$ prior factor, $k$-th \ac{LIO} relative pose factor, $l$-th \ac{VIO} relative pose factor, and $q$-th tilt prior factor, respectively.
Let $\mathcal{L}_k, \mathcal{V}_l$ be the relative pose measurement corresponding to the $k$-th \ac{LIO} and $l$-th \ac{VIO} relative pose factors, respectively.
The estimation problem can be equivalently expressed as a weighted nonlinear least squares problem as
\begin{multline}\label{eq:nls_graph}
  \footnotesize
  \hat{\mathcal{X}} = 
  \arg\min_\mathcal{X} \bigg(
  \sum_j \left|\left|
  \vec{e}^\mathrm{prior}_j(\mathcal{X}_j)
  \right|\right|_{\mat{\Sigma}_j^\mathrm{prior}}^2+\\
  +\sum_q \left|\left|
  \vec{e}^\mathrm{tilt}_q(\mathcal{X}_q)
  \right|\right|_{\mat{\Sigma}_q^\mathrm{tilt}}^2 +
  \sum_k \left|\left|
  \vec{e}^\mathrm{LIO}_k(\mathcal{X}_k,\mathcal{L}_k)
  \right|\right|_{\mat{\Sigma}_k^\mathrm{LIO}}^2+\\
  +\sum_l \left|\left|
  \vec{e}^\mathrm{VIO}_l(\mathcal{X}_l,\mathcal{V}_l)
  \right|\right|_{\mat{\Sigma}_l^\mathrm{VIO}}^2 +
  \sum_i \left|\left|
  \vec{e}^\mathrm{det}_i(\mathcal{X}_i,\vec{d}_i)
  \right|\right|_{\mat{\Sigma}_i^\mathrm{det}}^2
  \bigg),
\end{multline}
where $\vec{e}^\mathrm{prior}_j$ is the error function of the $SE(3)$ prior factors, $\vec{e}^\mathrm{tilt}_q$ is the error function of the zero-roll-pitch prior, $\vec{e}_k^\mathrm{LIO}$ is the error function of the \ac{LIO} relative poses of robot $X$, $\vec{e}_l^\mathrm{VIO}$ is the error function of the \ac{VIO} relative poses of robot $Y$, and $\vec{e}_i^\mathrm{det}$ is the error function of the 3D inter-robot detections between robot $X$ and robot $Y$.
Each error is weighted by the corresponding covariance matrix representing the Gaussian noise associated with the measurement.
The random $SE(3)$ variables~\cite{barfootAssociatingUncertaintyThreeDimensional2014} are defined with the tangent-space perturbations of the noise-free pose $\mat{T}$ applied on the right side:
\begin{equation}
  \tilde{\mat{T}} = \mat{T}\mathrm{Exp}(\vec{\xi}), \mat{T} \in SE(3), \vec{\xi} \sim \mathcal{N}(\vec{0}, \mat{\Sigma}),
\end{equation}
where $\mathrm{Exp()}$ is the exponential map of $SE(3)$, mapping elements of the vector space $\mathbb{R}^6$ to the Lie group $SE(3)$, and $\vec{\xi}$ is the perturbation vector in the tangent space with the convention of expressing orientation before translation to be compatible with the GTSAM library.
Assuming that $\mat{T}$ represents an $SE(3)$ pose in the world frame, the corresponding covariance matrix $\mat{\Sigma}$ is then defined in the body frame of $\mat{T}$.

To be able to incrementally build and solve the factor graph in real time, the graph is solved in a sliding window, utilizing a fixed-lag smoother based on the iSAM2~\cite{kaessISAM2IncrementalSmoothing2012}.
Due to the assumption that all the reference frames are gravity-aligned, we constrain the roll and pitch angles of all poses to zero using a prior factor on each estimated variable.

\subsection{LIO Relative Factor}

\noindent The relative pose is calculated from two consecutive outputs of the \ac{LIO} algorithm as
\begin{equation}
  \framess{X_k}{X_{k+1}}{\mat{T}} = 
  \left(\framess{L}{X_k}{\mat{T}}\right)^{-1}
  \framess{L}{X_{k+1}}{\mat{T}}.
\end{equation}
We assume that the \ac{LIO} utilizes a scan matching algorithm minimizing a point-to-line or point-to-plane cost function for matching the current \ac{lidar} scan to a map, and we utilize binary detection of \ac{LIO} degeneration based on the minimum eigenvalue of the approximate Hessian evaluated by the optimization method.
For a detailed description of the scan matching degeneracy detection, see~\cite{zhangDegeneracyOptimizationbasedState2016a}.
We consider the scan-to-map matching problem and the current \ac{LIO} output degenerated if
\begin{equation}
  \min \left(\mathrm{eig}\left(\mat{A}^\mathrm{T} \mat{A}\right)\right) < \lambda_\mathrm{thr},
\end{equation}
where $\mat{A}$ is the Jacobian of the scan-matching cost function evaluated at the current linearization point and $\lambda_\mathrm{thr}$ is a predefined threshold.
We assume that when the \ac{LIO} is not degenerated, it is more reliable than the \ac{VIO} method, and when the \ac{LIO} is degenerated, the \ac{LIO} position and yaw output are unusable.
Therefore, we set the covariance matrix of the \ac{LIO} relative pose as
\begin{equation}
  \mat{\Sigma}_k^\mathrm{\mathrm{L}IO} = \mathrm{diag}\left(\sigma^2_\alpha, \sigma^2_\beta, \prescript{\mathrm{L}}{}{\sigma}^2_\gamma, \prescript{\mathrm{L}}{}{\sigma}^2_\mathrm{pos}, \prescript{\mathrm{L}}{}{\sigma}^2_\mathrm{pos}, \prescript{\mathrm{L}}{}{\sigma}^2_\mathrm{pos}\right),
\end{equation}
where $\sigma^2_\alpha$ and $\sigma^2_\beta$ represent the variances of the relative roll and pitch angles, respectively, and are equal for both the reliable and degenerated \ac{LIO} case, as we constrain the roll and pitch angles to be constantly zero.
The yaw variance is obtained as
\begin{equation}
  \prescript{\mathrm{L}}{}{\sigma}^2_\gamma = \begin{cases}
    \Delta t \prescript{\mathrm{lo}}{}{\sigma}^2_\gamma, & \text{if } \framess{L}{X_k}{\mat{T}}, \framess{L}{X_{k+1}}{\mat{T}} \text{ are reliable} \\
    \Delta t \prescript{\mathrm{hi}}{}{\sigma}^2_\gamma, & \text{if either } \framess{L}{X_k}{\mat{T}}, \framess{L}{X_{k+1}}{\mat{T}} \text{ is degenerated}
  \end{cases},
\end{equation}
and the positional variances are calculated as
\begin{equation}
  \prescript{\mathrm{L}}{}{\sigma}^2_\mathrm{pos} = \begin{cases}
    \frac{\Delta t}{3}\prescript{\mathrm{lo}}{}{\sigma}^2_\mathrm{3D}, & \text{if } \framess{L}{X_k}{\mat{T}}, \framess{L}{X_{k+1}}{\mat{T}} \text{ are reliable} \\
    \frac{\Delta t}{3}\prescript{\mathrm{hi}}{}{\sigma}^2_\mathrm{3D}, & \text{if either } \framess{L}{X_k}{\mat{T}}, \framess{L}{X_{k+1}}{\mat{T}} \text{ is degenerated}
  \end{cases}.
\end{equation}
$\Delta t$ is the difference of timestamps of poses $\framess{L}{X_k}{\mat{T}}, \framess{L}{X_{k+1}}{\mat{T}}$, used to adapt the factor weighting to different sampling rates of the \ac{LIO} poses.
$\prescript{\mathrm{lo}}{}{\sigma}_\gamma, \prescript{\mathrm{hi}}{}{\sigma}_\gamma, \prescript{\mathrm{lo}}{}{\sigma}_\mathrm{3D}, \prescript{\mathrm{hi}}{}{\sigma}_\mathrm{3D}$ are predefined parameters representing the standard deviations of the yaw and 3D positional part of the \ac{LIO} relative pose in the reliable and degraded case, respectively.
The standard deviations are selected such that
\begin{equation}
  \prescript{\mathrm{lo}}{}{\sigma}_\gamma \ll \prescript{\mathrm{hi}}{}{\sigma}_\gamma,~~~\prescript{\mathrm{lo}}{}{\sigma}_\mathrm{3D} \ll \prescript{\mathrm{hi}}{}{\sigma}_\mathrm{3D}
\end{equation}
and based on analyzing the average error of the \ac{LIO} relative poses w.r.t. ground-truth measurements in real-world data.


\subsection{VIO Relative Factor}

\noindent The relative pose is calculated from two consecutive outputs of the \ac{VIO} algorithm as
\begin{equation}
  \framess{Y_l}{Y_{l+1}}{\mat{T}} = 
  \left(\framess{V}{Y_l}{\mat{T}}\right)^{-1}
  \framess{V}{Y_{l+1}}{\mat{T}}.
\end{equation}
We assume that the \ac{VIO} algorithm is based on a Kalman filter and encodes uncertainty information in the covariance matrix of each output.
The consecutive local poses are described by the jointly Gaussian probability distribution
\begin{equation}
  \mathcal{N}\left(\begin{bmatrix}\framess{V}{Y_l}{\mat{T}} \\ \framess{V}{Y_{l+1}}{\mat{T}}
  \end{bmatrix},
  \begin{bmatrix}
    \mat{\Sigma}_{l,l} & \mat{\Sigma}_{l,l+1} \\
    \mat{\Sigma}_{l+1,l} & \mat{\Sigma}_{l+1,l+1} \\
  \end{bmatrix}
    \right).
\end{equation}
Assuming that all the covariance matrices are defined in the same reference frame, the covariance matrix of the relative pose can be calculated as
\begin{equation}
  \mat{\Sigma}_\mathrm{rel} = \mat{\Sigma}_{l,l} + \mat{\Sigma}_{l+1,l+1} - \mat{\Sigma}_{l,l+1} - \mat{\Sigma}_{l+1,l},
\end{equation}
where $\mat{\Sigma}_{l,l}$ and $\mat{\Sigma}_{l+1,l+1}$ are the covariance matrices estimated by the Kalman filter at time steps $l$ and $l+1$, respectively.
The cross-covariance matrix $\mat{\Sigma}_{l,l+1}$ between the consecutive Kalman filter estimates is generally unknown, but ignoring the cross-covariance would lead to degradation of the estimate~\cite{mangelsonCharacterizingUncertaintyJointly2020}.
Therefore, we approximate the relative pose covariance matrix by a matrix proportional to the 2-Wasserstein distance~\cite{olkinDistanceTwoRandom1982, dowsonFrechetDistanceMultivariate1982a, bhatiaBuresWassersteinDistance2019} between the covariance matrices of the consecutive filter outputs.
For the covariance matrices $\mat{\Sigma}_{l,l}, \mat{\Sigma}_{l+1,l+1}$, the squared 2-Wasserstein distance between them is the solution of the problem
\begin{subequations}
\begin{equation}
  W_2\left(\mat{\Sigma}_{l,l}, \mat{\Sigma}_{l+1,l+1}\right)^2 = \min_{\mat{\Sigma}_{l,l+1}} \mathrm{Tr}\left(\mat{\Sigma}_{l,l} + \mat{\Sigma}_{l+1,l+1} - 2 \mat{\Sigma}_{l,l+1}\right)
\end{equation}
\begin{equation}
  \mathrm{s.t.}~
  \begin{bmatrix}
    \mat{\Sigma}_{l,l} & \mat{\Sigma}_{l,l+1} \\
    \mat{\Sigma}_{l+1,l} & \mat{\Sigma}_{l+1,l+1} \\
  \end{bmatrix}
  \ge 0.
\end{equation}
\end{subequations}
The Wasserstein distance is a statistical distance and a metric on the space of covariance matrices, and its calculation assumes maximal correlation between the two covariance matrices and thus provides a lower bound on the trace of the relative pose covariance matrix.
The squared 2-Wasserstein distance between the two Gaussians can be calculated as
\begin{multline}
  W_2\left(\mat{\Sigma}_{l,l},\mat{\Sigma}_{l+1,l+1}\right)^2 =\\
  \mathrm{Tr}\left[\mat{\Sigma}_{l,l} + \mat{\Sigma}_{l+1,l+1} - 2\left(\mat{\Sigma}_{l,l}^{\frac{1}{2}} \mat{\Sigma}_{l+1,l+1} \mat{\Sigma}_{l,l}^\frac{1}{2}\right)^\frac{1}{2}\right].
\end{multline}
To utilize this calculation, the covariance matrices $\mat{\Sigma}_{l,l}, \mat{\Sigma}_{l+1,l+1}$ need to be defined in the same reference frame.
In our case, the covariance matrices outputted by the \ac{VIO} algorithm are defined on the $SO(3) \times \mathbb{R}^3$ manifold.
In our approach, we utilize only the positional part of the covariance matrices; therefore, we can directly calculate the Wasserstein distance between the positional covariance matrices with no transformation necessary.
We assume that the error of the relative pose measurement is proportional to the Wasserstein distance between the two covariance matrices.
However, the Wasserstein distance provides only a lower bound on the trace of the relative pose covariance matrix, and using it directly as the trace of the relative pose covariance matrix would result in an overconfident estimate.
Furthermore, we do not assume that the covariance matrix always provides a consistent estimate of the uncertainty, i.e., the incoming covariance matrices $\mat{\Sigma}_{l,l}, \mat{\Sigma}_{l+1,l+1}$ may be underconfident or overconfident.
Therefore, we scale it by a constant factor as
\begin{equation}\label{eq:vio_sigma}
  \prescript{\mathrm{V}}{}{\sigma}^2_\mathrm{pos} = \frac{\mu}{3} W_2\left(\mat{\Sigma}_{l,l},\mat{\Sigma}_{l+1,l+1}\right)^2.
\end{equation}
The scaling factor $\mu$ depends on the system model of the \ac{VIO} algorithm and the odometry sampling rate and is selected based on analyzing the average error of the \ac{VIO} output with respect to ground-truth measurements on real-world data.
Next, we bound the variance to always stay between predefined thresholds as
\begin{equation}
  \prescript{\mathrm{V}}{}{\bar{\sigma}}^2_\mathrm{pos} = \max\left(\frac{\Delta t}{3} \prescript{\mathrm{Vmin}}{}{\sigma}^2_\mathrm{3D}, \min\left(\frac{\Delta t}{3} \prescript{\mathrm{Vmax}}{}{\sigma}^2_\mathrm{3D}, \prescript{\mathrm{V}}{}{\sigma}^2_\mathrm{3D}\right)\right).
\end{equation}
The thresholds are selected such that
\begin{equation}
\prescript{\mathrm{lo}}{}{\sigma}^2_\mathrm{3D} \leq \prescript{\mathrm{Vmin}}{}{\sigma}^2_\mathrm{3D} < \prescript{\mathrm{Vmax}}{}{\sigma}^2_\mathrm{3D} \leq \prescript{\mathrm{hi}}{}{\sigma}^2_\mathrm{3D}.
\end{equation}
The variance of the yaw measurement is calculated as
\begin{equation}\label{eq:vio_yaw_inflation}
  \prescript{\mathrm{V}}{}{\bar{\sigma}}^2_\gamma = \begin{cases}
    \Delta t \prescript{\mathrm{V}}{}{\sigma}^2_\gamma & \text{if } \prescript{\mathrm{V}}{}{\sigma}^2_\mathrm{pos} \leq \frac{\Delta t}{3} \prescript{\mathrm{Vmax}}{}{\sigma}^2_\mathrm{3D} \\
    \nu \Delta t \prescript{\mathrm{V}}{}{\sigma}^2_\gamma & \text{if }  \prescript{\mathrm{V}}{}{\sigma}^2_\mathrm{pos} > \frac{\Delta t}{3} \prescript{\mathrm{Vmax}}{}{\sigma}^2_\mathrm{3D}
  \end{cases},
\end{equation}
where $\prescript{\mathrm{V}}{}{\sigma}_\gamma$ is an empirically-selected parameter and $\nu$ is a predefined scaling factor inflating the yaw sigma if the positional measurement is deemed unreliable due to high positional standard deviation.
The covariance matrix of the relative pose measurement is set as
\begin{equation}
  \mat{\Sigma}_l^\mathrm{VIO} = \mathrm{diag}\left(\sigma_\alpha^2, \sigma_\beta^2, \prescript{\mathrm{V}}{}{\bar{\sigma}}_\gamma^2, \prescript{\mathrm{V}}{}{\bar{\sigma}}_\mathrm{pos}^2, \prescript{\mathrm{V}}{}{\bar{\sigma}}_\mathrm{pos}^2, \prescript{\mathrm{V}}{}{\bar{\sigma}}_\mathrm{pos}^2\right).
\end{equation}


\subsection{Interpolation-Based Detection Factor}
\noindent The relative robot detection is represented by a quaternary factor connecting the temporally-adjacent variables $\framess{W}{X_k}{\mat{T}}, \framess{W}{X_{k+1}}{\mat{T}}, \framess{W}{Y_l}{\mat{T}}, \framess{W}{Y_{l+1}}{\mat{T}}$.
The factor utilizes the assumption of constant velocity between consecutive variables.
The error function is defined as
\begin{equation}\label{eq:measurement_dist}
  \vec{e}_i^\mathrm{det} =
  \left[ \left(\framess{W}{X_\mathrm{int}}{\mat{T}}\right)^{-1}
  \framess{W}{Y_\mathrm{int}}{\mat{T}} \right]_\mathrm{tr} - \vec{d}_i, 
\end{equation}
where $\vec{d}$ is the detection, $[~]_\mathrm{tr}$ selects the translational part of the $\mathrm{SE(3)}$ matrix, and $\framess{W}{X_\mathrm{int}}{\mat{T}}$, $\framess{W}{Y_\mathrm{int}}{\mat{T}}$ are obtained by linear interpolation on the $\mathrm{SE(3)}$ manifold as
\begin{subequations}
\begin{equation}
\framess{W}{X_\mathrm{int}}{\mat{T}} = 
  \framess{W}{X_k}{\mat{T}} \mathrm{Exp}\left(\tau_X\mathrm{Log}\left( \left(\framess{W}{X_k}{\mat{T}}\right)^{-1} \framess{W}{X_{k+1}}{\mat{T}} \right) \right),
\end{equation}
\begin{equation}
\framess{W}{Y_\mathrm{int}}{\mat{T}} = 
  \framess{W}{Y_l}{\mat{T}} \mathrm{Exp}\left(\tau_Y\mathrm{Log}\left( \left(\framess{W}{Y_l}{\mat{T}}\right)^{-1} \framess{W}{Y_{l+1}}{\mat{T}} \right) \right),
\end{equation}
\begin{equation}
  \tau_X = \frac{t_\mathrm{det} - t_{X_k}}{t_{X_{k+1}} - t_{X_k}};~
  \tau_Y = \frac{t_\mathrm{det} - t_{Y_l}}{t_{Y_{l+1}} - t_{Y_l}},
\end{equation}
\end{subequations}
where $t_\mathrm{det}$ is the time of the detection, $t_A$ represents time of variable $\framess{W}{A}{\mat{T}}$, $\mathrm{Exp}$ denotes the $SE(3)$ exponential map, and $\mathrm{Log}$ is the $SE(3)$ logarithm map.

Let us denote $\vec{\xi}_A$ the perturbation vector on the tangent space associated with the pose $\framess{W}{A}{\mat{T}}$.
The analytic Jacobian matrix of the detection factor error function at $\framess{W}{X_k}{\mat{T}}, \framess{W}{X_{k+1}}{\mat{T}}, \framess{W}{Y_l}{\mat{T}}, \framess{W}{Y_{l+1}}{\mat{T}}$  is formulated as
\begin{equation}
  \mat{J} = \begin{bmatrix}
    \frac{\partial\vec{e}_\mathrm{det}}{\partial \vec{\xi}_{X_k}} &

    \frac{\partial\vec{e}_\mathrm{det}}{\partial \vec{\xi}_{X_{k+1}}} &

    \frac{\partial\vec{e}_\mathrm{det}}{\partial \vec{\xi}_{Y_l}} &

    \frac{\partial\vec{e}_\mathrm{det}}{\partial \vec{\xi}_{Y_{l+1}}}
  \end{bmatrix},
\end{equation}
where the submatrices are calculated using the chain rule as
\begin{subequations}
\begin{align}
    \frac{\partial\vec{e}_\mathrm{det}}{\partial \vec{\xi}_{X_k}} = 
    \frac{\partial\vec{e}_\mathrm{det}}{\partial \vec{\xi}_{X_\mathrm{int}}}
    \frac{\partial\framess{ }{X_\mathrm{int}}{\vec{\xi}}}{\partial \vec{\xi}_{X_k}} &,~
    \frac{\partial\vec{e}_\mathrm{det}}{\partial \vec{\xi}_{X_{k+1}}} = 
    \frac{\partial\vec{e}_\mathrm{det}}{\partial \vec{\xi}_{X_\mathrm{int}}}
    \frac{\partial\framess{ }{X_\mathrm{int}}{\vec{\xi}}}{\partial \vec{\xi}_{X_{k+1}}},\\
    \frac{\partial\vec{e}_\mathrm{det}}{\partial \vec{\xi}_{Y_l}} = 
    \frac{\partial\vec{e}_\mathrm{det}}{\partial \vec{\xi}_{Y_\mathrm{int}}}
    \frac{\partial\framess{ }{Y_\mathrm{int}}{\vec{\xi}}}{\partial \vec{\xi}_{Y_l}} &,~
    \frac{\partial\vec{e}_\mathrm{det}}{\partial \vec{\xi}_{Y_{l+1}}} = 
    \frac{\partial\vec{e}_\mathrm{det}}{\partial \vec{\xi}_{Y_\mathrm{int}}}
    \frac{\partial\framess{ }{Y_\mathrm{int}}{\vec{\xi}}}{\partial \vec{\xi}_{Y_{l+1}}}.
\end{align}
\end{subequations}
  The partial derivatives of the detection error function with respect to perturbations of the interpolated poses $\framess{W}{X_\mathrm{int}}{\mat{T}}$ and $\framess{W}{Y_\mathrm{int}}{\mat{T}}$ are formulated as
\begin{subequations}
\begin{equation}
  \frac{\partial\vec{e}_\mathrm{det}}{\partial \vec{\xi}_{X_\mathrm{int}}}
  = \begin{bmatrix}
    \left(\framess{W}{X_\mathrm{int}}{\mat{R}} \right)^{\mathrm{T}}
    \left[ \framess{W}{Y_\mathrm{int}}{\vec{t}} - \framess{W}{X_\mathrm{int}}{\vec{t}} \right]_{\times} \framess{W}{X_\mathrm{int}}{\mat{R}} 
    & -\mat{I}_3
  \end{bmatrix},
\end{equation}
\begin{equation}
  \frac{\partial\vec{e}_\mathrm{det}}{\partial \vec{\xi}_{Y_\mathrm{int}}}
  = \begin{bmatrix}
    \mat{0}_3 &
\left(\framess{W}{X_\mathrm{int}}{\mat{R}} \right)^{\mathrm{T}}
\framess{W}{Y_\mathrm{int}}{\mat{R}}
  \end{bmatrix},
\end{equation}
\end{subequations}
where $[\vec{a}]_\times$ denotes a skew-symmetric matrix constructed from vector $\vec{a}$.
The partial derivatives of the interpolated poses with respect to perturbations of the estimated variables are
\begin{subequations}
\begin{equation}
  \frac{\partial\framess{ }{A_\mathrm{int}}{\vec{\xi}}}{\partial \vec{\xi}_{A_k}}
  = \mat{Ad}_{\framess{A_\mathrm{int}}{A_k}{\mat{T}}} - \tau_A \mat{J}^{A}_\mathrm{Exp} \mat{J}^A_\mathrm{Log} \mat{Ad}_{\framess{A_{k+1}}{W}{\mat{T}} \framess{W}{A_k}{\mat{T}} },
\end{equation}
\begin{align}
  \frac{\partial\framess{ }{A_\mathrm{int}}{\vec{\xi}}}{\partial \vec{\xi}_{A_{k+1}}} 
  = \tau_A \mat{J}^{A}_\mathrm{Exp} \mat{J}^A_\mathrm{Log},
\end{align}
\end{subequations}
where $\mat{Ad_T}$ denotes the adjoint of $\mat{T}$ and the specific partial derivative can be obtained by substituting $A_k, A_{k+1}$ with $X_k, X_{k+1}$ or $Y_l, Y_{l+1}$.
The Jacobians of the exponential and logarithm maps are formulated as
\begin{subequations}
\begin{equation}
  \mat{J}^{A}_\mathrm{Exp} = \mat{J}_r\big|_{\tau_A\mathrm{Log}\left( \left(\framess{W}{A_k}{\mat{T}} \right)^{-1} \framess{W}{A_{k+1}}{\mat{T}} \right)},
\end{equation}
\begin{equation}
  \mat{J}^A_\mathrm{Log} = \mat{J}_r^{-1}\big|_{ \left(\framess{W}{A_k}{\mat{T}} \right)^{-1} \framess{W}{A_{k+1}}{\mat{T}}},
\end{equation}
\end{subequations}
where $\mat{J}_r$, $\mat{J}_r^{-1}$ denote the right Jacobian of $SE(3)$ and its inverse, evaluated at the corresponding linearization points~\cite{solaMicroLieTheory2021}.
The detection covariance matrix $\mat{\Sigma}^\mathrm{det}$ is set as
\begin{equation}
  \mat{\Sigma}_i^\mathrm{det} = \mathrm{diag}\left(\sigma^2_\mathrm{det}, \sigma^2_\mathrm{det}, \sigma^2_\mathrm{det}\right)
\end{equation}
where $\sigma_\mathrm{det}$ are constant values, which are empirically tuned on real-world data to reflect the noise of the detections and the inaccuracies caused by the constant-velocity assumption used by the factor.

The detections are anonymous, i.e., the detections contain no information about which robot is detected and may contain false positives.
Therefore, it is necessary to associate each incoming detection to a tracked robot before the detection is inserted into the graph.
To perform the association, the measurement distance of the detection is calculated based on eq.~(\ref{eq:measurement_dist}) and the detection is associated to the closest robot if the measurement distance is below a predefined threshold.

\subsection{Graph Initialization}
\noindent To initialize the factor graph, we need an initial guess for the pose of each robot.
We assume that there is no prior information about the position and yaw orientation of the robots.
The detections are anonymous, i.e., the detections contain no information about which robot is detected and may contain false positives.
Therefore, we obtain the initial pose of each robot by aligning its movement trajectory observed in the detections and in the odometry data.

We assume that the detector provides tracking of the detected object, i.e., we can construct a buffer of detections corresponding to a single object, but it is unknown which buffer corresponds to which robot.
For each robot and for each detection buffer, we construct a sliding window of detections and corresponding robot poses, and solve the problem of alignment of the two point patterns as
\begin{subequations}
\begin{equation}\label{eq:init}
  \framess{L}{V}{\hat{\vec{t}}}, \hat{\theta}= \arg\min_{\framess{L}{V}{\vec{t}}, \theta} \sum_i\left|\left|\framess{L}{V}{\mat{R}}(\theta)\framess{V}{Y_i}{\vec{p}} + \framess{L}{V}{\vec{t}} - \framess{L}{Y_i}{\vec{d}}\right|\right|^2,
\end{equation}
\begin{equation}
  \framess{L}{V}{\vec{t}} \in \mathbb{R}^3,~\theta\in [-\pi,\pi],
\end{equation}
\begin{equation}
  \framess{L}{V}{\mat{T}} = \begin{bmatrix}
    \mat{R}_z(\theta) & \framess{L}{V}{\vec{t}}\\
    \mat{0}^\mathrm{T} & 1\\
  \end{bmatrix} \in SE(3).
\end{equation}
\end{subequations}
To obtain correspondences at the same time steps, $\framess{V}{Y_i}{\vec{p}}$ is calculated by linear interpolation of the temporally-adjacent positions to the time of the detection $\framess{L}{Y_i}{\vec{d}}$.
The problem can be solved analytically using the Kabsch-Umeyama algorithm~\cite{umeyamaLeastsquaresEstimationTransformation1991}, solving only for the rotation and translation.
The initialization is considered successful if the cost of~(\ref{eq:init}) falls below a predefined threshold and the trajectories are sufficiently large, as the alignment problem degrades as the robot trajectories degrade to a single point, and noise in the data might compromise the result.

To initialize the factor graph, we anchor the pose of the robot $X$ at the origin
\begin{equation}
  \framess{W}{X_0}{\mat{T}} = \mat{I}_4
\end{equation}
and calculate prior poses of the remaining robots based on the transformations obtained from trajectory alignment as
\begin{equation}
  \framess{W}{Y_0}{\mat{T}} = \framess{L}{V}{\mat{T}} \framess{V}{Y_0}{\mat{T}}.
\end{equation}
The covariance matrices of the prior poses are empirically selected to reflect a strong constraint on anchoring $\framess{W}{X_0}{\mat{T}}$ at origin and a comparatively larger uncertainty of the prior pose $\framess{W}{Y_0}{\mat{T}}$, which can be corrected over the course of the estimation process.

To perform the initialization, each detected robot needs to separately perform an initialization maneuver so that the trajectory alignment problem can be solved.
As the estimated transformation $\framess{L}{V}{\mat{T}}$ has only 4 \acp{DOF}, this initialization maneuver can be a short flight in a straight line.

\section{Observability Analysis}
\label{sec:observability}

\begin{figure}[t]
  \centering
  \input{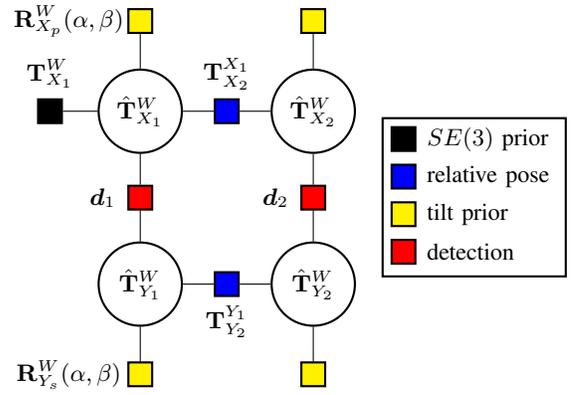}
  \caption{Simplified factor graph used in the observability analysis of the cooperative localization problem.
  The graph represents two poses of robot $X$ and two poses of robot $Y$.
  Detections are synchronized with the odometry measurements, removing the need for interpolation.
  In the visualized situation, the factor graph is fully constrained.
  }

  \label{fig:factor_graph_simple}
\end{figure}

\noindent To analyze the observability of the estimation problem with respect to various situations and sensory degradations, we construct the Jacobian matrix of the stacked measurement model of a simplified factor graph representing two poses of robot $X$ and two poses of robot $Y$ (see Fig.~\ref{fig:factor_graph_simple}).
By analyzing the rank of the resulting Jacobian matrix, we can determine whether the stacked measurements are sufficient for fully constraining the estimated variables.
Furthermore, by analyzing the nullspace of the Jacobian matrix, we can obtain the unobservable directions, i.e., the subspace of the values of the estimated variables whose changes result in no change of the cost function.

For clarity of the analysis, we consider robot detections to be synchronized with the odometry measurements, thus removing the need for pose interpolation, and we assume that the roll and pitch angles of all poses are zero, with the orientation depending on yaw only.
We construct the following Jacobian matrices from partial derivatives of the cost function~(\ref{eq:nls_graph}) w.r.t. estimated robot poses:
\begin{subequations}
\begin{equation}
  \mat{J}^\mathrm{LIO} = \begin{bmatrix}
    \frac{\partial\vec{e}^\mathrm{LIO}}{\partial \vec{\xi}_{X_\mathrm{1}}} & \frac{\partial\vec{e}^\mathrm{LIO}}{\partial \vec{\xi}_{X_\mathrm{2}}} & \mat{0}_{6\times6} & \mat{0}_{6\times6} \\
  \end{bmatrix} \in \mathbb{R}^{6\times24},
\end{equation}
\begin{equation}
  \mat{J}^\mathrm{VIO} = \begin{bmatrix}
    \mat{0}_{6\times6} & \mat{0}_{6\times6} & \frac{\partial\vec{e}^\mathrm{VIO}}{\partial \vec{\xi}_{Y_\mathrm{1}}} & \frac{\partial\vec{e}^\mathrm{VIO}}{\partial \vec{\xi}_{Y_\mathrm{2}}} \\
  \end{bmatrix} \in \mathbb{R}^{6\times24},
\end{equation}
\begin{equation}
  \mat{J}^\mathrm{det}_1 = \begin{bmatrix}
    \frac{\partial\vec{e}^\mathrm{det}_1}{\partial \vec{\xi}_{X_\mathrm{1}}} & \mat{0}_{3\times6} & \frac{\partial\vec{e}^\mathrm{det}_1}{\partial \vec{\xi}_{Y_\mathrm{1}}} & \mat{0}_{3\times6} \\
  \end{bmatrix} \in \mathbb{R}^{3\times24},
\end{equation}
\begin{equation}
  \mat{J}^\mathrm{det}_2 = \begin{bmatrix}
    \mat{0}_{3\times6} & \frac{\partial\vec{e}^\mathrm{det}_2}{\partial \vec{\xi}_{X_\mathrm{2}}} & \mat{0}_{3\times6} & \frac{\partial\vec{e}^\mathrm{det}_2}{\partial \vec{\xi}_{Y_\mathrm{2}}} \\
  \end{bmatrix} \in \mathbb{R}^{3\times24},
\end{equation}
\begin{equation}
  \mat{J}^\mathrm{prior}_X = \begin{bmatrix}
    \mat{I}_{6\times6} & \mat{0}_{6\times6} & \mat{0}_{6\times6} & \mat{0}_{6\times6} \\
  \end{bmatrix} \in \mathbb{R}^{6\times24},
\end{equation}
\begin{equation}
  \mat{J}^\mathrm{prior}_Y = \begin{bmatrix}
    \mat{0}_{6\times6} & \mat{I}_{6\times6} & \mat{0}_{6\times6} & \mat{0}_{6\times6} \\
  \end{bmatrix} \in \mathbb{R}^{6\times24},
\end{equation}
\begin{equation}
  \mat{J}_\mathrm{tilt} = \begin{bmatrix}
    \mat{I}_{2\times 2} & \mat{0}_{2\times 4}
  \end{bmatrix} \in \mathbb{R}^{2\times6},
\end{equation}
\begin{equation}
  \mat{J}^\mathrm{tilt}_4 = \begin{bmatrix}
    \mat{J}_\mathrm{tilt} & \mat{0}_{2\times6} & \mat{0}_{2\times6} & \mat{0}_{2\times6} \\
    \mat{0}_{2\times6} & \mat{J}_\mathrm{tilt} & \mat{0}_{2\times6} & \mat{0}_{2\times6} \\
    \mat{0}_{2\times6} & \mat{0}_{2\times6} & \mat{J}_\mathrm{tilt} & \mat{0}_{2\times6} \\
    \mat{0}_{2\times6} & \mat{0}_{2\times6} & \mat{0}_{2\times6} & \mat{J}_\mathrm{tilt} \\
  \end{bmatrix} \in \mathbb{R}^{8\times24},
\end{equation}
\end{subequations}
corresponding to Jacobian of the \ac{LIO} odometry factor, \ac{VIO} odometry factor, detection between $\framess{W}{X_1}{\mat{T}}$ and $\framess{W}{Y_1}{\mat{T}}$, detection between $\framess{W}{X_2}{\mat{T}}$ and $\framess{W}{Y_2}{\mat{T}}$, $SE(3)$ priors on poses $\framess{W}{X_1}{\mat{T}}$ and $\framess{W}{Y_1}{\mat{T}}$, and prior on the roll and pitch angles of all poses, respectively.
The values of the partial derivatives are formulated as
\begin{subequations}
  \label{eq:partial}
\begin{equation}
  \label{eq:partial1}
  \frac{\partial\vec{e}^\mathrm{LIO}}{\partial \vec{\xi}_{X_\mathrm{1}}} =
  \begin{bmatrix}
    - \left(\framess{W}{X_2}{\mat{R}}\right)^\mathrm{T} \framess{W}{X_1}{\mat{R}} & \mat{0}_{3\times 3} \\
    \left(\framess{W}{X_2}{\mat{R}}\right)^\mathrm{T} \left[ \framess{W}{X_2}{\vec{t}} - \framess{W}{X_1}{\vec{t}} \right]_\times \framess{W}{X_1}{\mat{R}} & - \left(\framess{W}{X_2}{\mat{R}}\right)^\mathrm{T} \framess{W}{X_1}{\mat{R}}
  \end{bmatrix}
\end{equation}
\begin{equation}
  \label{eq:partial2}
  \frac{\partial\vec{e}^\mathrm{LIO}}{\partial \vec{\xi}_{X_\mathrm{2}}} = \mat{I}_{6\times 6}
\end{equation}
\begin{equation}
  \label{eq:partial3}
\frac{\partial\vec{e}^\mathrm{det}_1}{\partial \vec{\xi}_{X_\mathrm{1}}}
  = \begin{bmatrix}
    \left(\framess{W}{X_\mathrm{1}}{\mat{R}} \right)^{\mathrm{T}}
    \left[ \framess{W}{Y_\mathrm{1}}{\vec{t}} - \framess{W}{X_\mathrm{1}}{\vec{t}} \right]_{\times} \framess{W}{X_\mathrm{1}}{\mat{R}} 
    & -\mat{I}_3
  \end{bmatrix},
\end{equation}
\begin{equation}
  \label{eq:partial4}
\frac{\partial\vec{e}^\mathrm{det}_1}{\partial \vec{\xi}_{Y_\mathrm{1}}}
  = \begin{bmatrix}
    \mat{0}_3 &
\left(\framess{W}{X_\mathrm{1}}{\mat{R}} \right)^{\mathrm{T}}
\framess{W}{Y_\mathrm{1}}{\mat{R}}
  \end{bmatrix},
\end{equation}
\end{subequations}
with $\frac{\partial\vec{e}^\mathrm{VIO}}{\partial \vec{\xi}_{Y_\mathrm{1}}}$, $\frac{\partial\vec{e}^\mathrm{VIO}}{\partial \vec{\xi}_{Y_\mathrm{2}}}$, $\frac{\partial\vec{e}^\mathrm{det}_2}{\partial \vec{\xi}_{X_\mathrm{2}}}$, and $\frac{\partial\vec{e}^\mathrm{det}_2}{\partial \vec{\xi}_{Y_\mathrm{2}}}$ formulated analogously by substituting the corresponding variables into the partial derivative equations.

\subsection{Full-Rank Jacobian}
\noindent The Jacobian of the full factor graph with both odometry measurements, both detections, prior on pose $\framess{W}{X_1}{\mat{T}}$, and prior on the roll and pitch angles is
\begin{equation}
  \mat{J}^\mathrm{full} = \begin{bmatrix}
    \mat{J}^\mathrm{LIO}, \mat{J}^\mathrm{VIO}, \mat{J}^\mathrm{det}_1, \mat{J}^\mathrm{det}_2, \mat{J}^\mathrm{prior}_X, \mat{J}^\mathrm{tilt}_4
  \end{bmatrix}^\mathrm{T}
\end{equation}
and its rank is
\begin{equation}
  \mathrm{rank}\left(\mat{J}^\mathrm{full}\right) = 24,
\end{equation}
demonstrating full observability of the estimation problem.

\subsection{No Global $SE(3)$ Prior}
\noindent By removing the $SE(3)$ prior on pose $\framess{W}{X_1}{\mat{T}}$, we obtain
\begin{subequations}
\begin{equation}
  \mat{J}^\mathrm{no}_\mathrm{prior} = \begin{bmatrix}
    \mat{J}^\mathrm{LIO}, \mat{J}^\mathrm{VIO}, \mat{J}^\mathrm{det}_1, \mat{J}^\mathrm{det}_2, \mat{J}^\mathrm{tilt}_4
  \end{bmatrix}^\mathrm{T},
\end{equation}
\begin{equation}
  \mathrm{rank}\left(\mat{J}^\mathrm{no}_\mathrm{prior}\right) = 20,
\end{equation}
\end{subequations}
exhibiting four unobservable \acp{DOF} corresponding to the translation and yaw orientation w.r.t. the global frame of reference.
As the graph does not contain any global measurements and the $SE(3)$ prior is utilized only to anchor the graph at an initial pose and to make the problem solvable, over time, the estimated poses will exhibit long-term drift with respect to the global reference frame.

\subsection{No Roll and Pitch Prior}
\noindent By removing the roll and pitch prior from Jacobian $\mat{J}_\mathrm{full}$, we obtain
\begin{subequations}
\begin{equation}
  \mat{J}^\mathrm{no}_\mathrm{rpprior} = \begin{bmatrix}
    \mat{J}^\mathrm{LIO}, \mat{J}^\mathrm{VIO}, \mat{J}^\mathrm{det}_1, \mat{J}^\mathrm{det}_2, \mat{J}^\mathrm{prior}_X
  \end{bmatrix}^\mathrm{T},
\end{equation}
\begin{equation}
  \mathrm{rank}\left(\mat{J}^\mathrm{no}_\mathrm{rpprior}\right) = 23.
\end{equation}
\end{subequations}
To obtain the unobservable direction, we calculate the nullspace of the Jacobian as
\begin{equation}
  \mathrm{Null}\left(\mat{J}^\mathrm{no}_\mathrm{rpprior}\right) = \mathrm{span}\left\{\begin{bmatrix}
    \mat{0}_\mathrm{12\times 1}\\
    \framess{Y_1}{W}{\mat{R}} \left(\framess{W}{Y_1}{\vec{t}} - \framess{W}{Y_2}{\vec{t}}\right) \\
    \mat{0}_\mathrm{3\times 1}\\
    \framess{Y_2}{W}{\mat{R}} \left(\framess{W}{Y_1}{\vec{t}} - \framess{W}{Y_2}{\vec{t}}\right) \\
    \mat{0}_{3\times 1}
  \end{bmatrix}\right\},
\end{equation}
showing that when robot $Y$ is moving in a straight line, its 3D orientation cannot be fully determined without additional information.
By incorporating the assumption of gravity-alignment of all reference frames in the form of the roll-pitch prior, our approach circumvents this unobservability.
Note that the unobservable direction is expressed in the body frame of the robot poses.

\subsection{No Movement of Robot $Y$}
\noindent By setting the pose $\framess{W}{Y_2}{\mat{T}}$ equal to pose $\framess{W}{Y_1}{\mat{T}}$, we obtain a situation with no movement of the robot $Y$:
\begin{subequations}
\begin{equation}
  \mat{J}^\mathrm{no}_\mathrm{move} = {\mat{J}^\mathrm{full}}\big|_{\framess{W}{Y_1}{\mat{T}} = \framess{W}{Y_2}{\mat{T}}},
\end{equation}
\begin{equation}
  \mathrm{rank}\left(\mat{J}^\mathrm{no}_\mathrm{move}\right) = 23,
\end{equation}
\begin{equation}
  \mathrm{Null}\left(\mat{J}^\mathrm{no}_\mathrm{move}\right) = \mathrm{span}\left\{ \begin{bmatrix}
    \mat{0}_\mathrm{14\times 1}\\
    1\\
    \mat{0}_\mathrm{5\times 1}\\
    1 \\
    \mat{0}_{3\times 1}
  \end{bmatrix}\right\},
\end{equation}
\end{subequations}
showing that without the movement of robot $Y$, we cannot fully determine the orientation of robot $Y$, as the relative orientation between the odometry frame of robot $Y$ and the odometry frame of robot $X$ is unobservable.

\subsection{LIO Degradation}
\label{sec:obs_lio}
\noindent In case of a detected \ac{LIO} degradation, the $\Sigma^\mathrm{LIO}$ covariance in eq.~(\ref{eq:nls_graph}) will be greatly increased, minimizing the influence of the \ac{LIO} term on the objective function.
Therefore, we remove the corresponding line from the Jacobian $\mat{J}_\mathrm{full}$.
By removing the part corresponding to the \ac{LIO} measurement from the full Jacobian and analyzing the result, we can ascertain which subspace of the estimated variables is not constrained by other measurements and thus cannot be corrected in the event of \ac{LIO} degradation.
We assume that the degradation happens after a period of normal operation, and the graph contains valid prior estimates of both robot poses.
We construct the Jacobian of a factor graph with \ac{LIO} degradation as
\begin{subequations}
\begin{equation}
  \mat{J}^\mathrm{LIO}_\mathrm{deg} = \begin{bmatrix}
    \mat{J}^\mathrm{VIO}, \mat{J}^\mathrm{det}_1, \mat{J}^\mathrm{det}_2, \mat{J}^\mathrm{prior}_X, \mat{J}^\mathrm{prior}_Y, \mat{J}^\mathrm{tilt}_4
  \end{bmatrix}^\mathrm{T},
\end{equation}
\begin{equation}
  \mathrm{rank}\left(\mat{J}^\mathrm{LIO}_\mathrm{deg}\right) = 23,
\end{equation}
\begin{equation}
  \mathrm{Null}\left(\mat{J}^\mathrm{LIO}_\mathrm{deg}\right) = \mathrm{span} \left\{ \begin{bmatrix}
    \mat{0}_\mathrm{8\times 1}\\
    1 \\
    \left[\framess{X_2}{W}{\mat{R}} \left(\framess{W}{X_2}{\vec{t}} - \framess{W}{Y_2}{\vec{t}}\right)^\bot\right]_{x} \\
    \left[\framess{X_2}{W}{\mat{R}} \left(\framess{W}{X_2}{\vec{t}} - \framess{W}{Y_2}{\vec{t}}\right)^\bot\right]_{y} \\
    \mat{0}_{13\times 1}
  \end{bmatrix} \right\},
\end{equation}
\end{subequations}
indicating that in such a case, there is insufficient information to distinguish between the change in the yaw of the robot $X$ and its movement in the direction perpendicular to the line connecting the robots.
The notation $\vec{x}^\bot$ represents a vector perpendicular to vector $\vec{x}$ and the subscripts $[\vec{a}]_x$ and $[\vec{a}]_y$ represent the $x$- or $y$-component of vector $\vec{a}$.
By inserting an additional detected robot $Z$, we obtain
\begin{subequations}
\begin{equation}
  \mat{J}^{\mathrm{LIO}\prime}_\mathrm{deg} = \begin{bmatrix}
    \multicolumn{4}{c}{\mat{J}^\mathrm{LIO}_\mathrm{deg}} &  \mat{0}_{32\times6} & \mat{0}_{32\times6} \\

    \mat{0}_{6\times6} & \mat{0}_{6\times6} & \mat{0}_{6\times6} & \mat{0}_{6\times6} & \frac{\partial\vec{e}^\mathrm{VIO}}{\partial \vec{\xi}_{Z_\mathrm{1}}} & \frac{\partial\vec{e}^\mathrm{VIO}}{\partial \vec{\xi}_{Z_\mathrm{2}}} \\

    \frac{\partial\vec{e}^\mathrm{det}_3}{\partial \vec{\xi}_{X_\mathrm{1}}} & \mat{0}_{3\times6} & \mat{0}_{3\times6} & \mat{0}_{3\times6} & \frac{\partial\vec{e}^\mathrm{det}_3}{\partial \vec{\xi}_{Z_\mathrm{1}}} & \mat{0}_{3\times6}\\

    \mat{0}_{3\times6} & \frac{\partial\vec{e}^\mathrm{det}_4}{\partial \vec{\xi}_{X_\mathrm{2}}} & \mat{0}_{3\times6} & \mat{0}_{3\times6} & \mat{0}_{3\times6} & \frac{\partial\vec{e}^\mathrm{det}_4}{\partial \vec{\xi}_{Z_\mathrm{2}}}\\

    \mat{0}_{6\times6} & \mat{0}_{6\times6} & \mat{0}_{6\times6} & \mat{0}_{6\times6} & \mat{I}_{6\times6} & \mat{0}_{6\times6} \\

    \mat{0}_{2\times6} & \mat{0}_{2\times6} & \mat{0}_{2\times6} & \mat{0}_{2\times6} & \mat{J}_\mathrm{tilt} & \mat{0}_{2\times6} \\
    \mat{0}_{2\times6} & \mat{0}_{2\times6} & \mat{0}_{2\times6} & \mat{0}_{2\times6} & \mat{0}_{2\times6} & \mat{J}_\mathrm{tilt} \\

  \end{bmatrix},
\end{equation}
\begin{equation}
  \mathrm{rank}\left(\mat{J}^{\mathrm{LIO}\prime}_\mathrm{deg}\right) = 36.
\end{equation}
\end{subequations}
The resulting Jacobian is full-rank, showing that the problem becomes fully observable when two distinct robots with reliable pose estimates are detected.

\subsection{VIO Degradation}
\label{sec:obs_vio}
\noindent In case of a \ac{VIO} degradation, the influence of the VIO term from eq.~(\ref{eq:nls_graph}) is minimized.
Therefore, we obtain
\begin{subequations}
\begin{equation}
  \mat{J}^\mathrm{VIO}_\mathrm{deg} = \begin{bmatrix}
    \mat{J}^\mathrm{LIO}, \mat{J}^\mathrm{det}_1, \mat{J}^\mathrm{det}_2, \mat{J}^\mathrm{prior}_X, \mat{J}^\mathrm{prior}_Y, \mat{J}^\mathrm{tilt}_4
  \end{bmatrix}^\mathrm{T},
\end{equation}
\begin{equation}
  \mathrm{rank}\left(\mat{J}^\mathrm{VIO}_\mathrm{deg}\right) = 23,
\end{equation}
\begin{equation}
  \mathrm{Null}\left(\mat{J}^\mathrm{VIO}_\mathrm{deg}\right) = \mathrm{span} \left\{ \begin{bmatrix}
    \mat{0}_\mathrm{20\times 1}\\
    1 \\
    \mat{0}_{3\times 1}
  \end{bmatrix} \right\},
\end{equation}
\end{subequations}
indicating that the yaw of pose $\framess{W}{Y_2}{\mat{T}}$ is unobservable from the available measurements.
Such a situation cannot be improved by adding additional robots, as the detections do not provide any orientation information about the robot $Y$, and the robot $Y$ itself cannot detect other robots.



\section{Experimental Verification}
\label{sec:experiments}

\begin{table}[t]
\small
\begin{center}
  \caption{Parameter values used in the experiments. All the presented results were obtained with the same set of parameters. $\mu_Y$ and $\mu_Z$ represent different values of $\mu$ for \acp{UAV} $Y$ and $Z$ selected due to the \acp{UAV}' different construction resulting in different odometry accuracy.}
  \begin{tabularx}{1.0\linewidth}{X X} 
    \toprule 
    \textbf{Parameter} & \textbf{Value} \\ \midrule
    odometry rate & \SI{2}{Hz} \\
    detection rate & \SI{10}{Hz} \\
    smoother window length & \SI{30}{s} \\

    $\lambda_\mathrm{thr}$ & \num{430} \\

    $\sigma_\mathrm{det}$ &  \SI{0.13}{m} \\
    $\prescript{\mathrm{lo}}{}{\sigma}_\mathrm{3D}$ & \SI{0.01}{m} \\
    $\prescript{\mathrm{hi}}{}{\sigma}_\mathrm{3D}$ & \SI{5.0}{m} \\
    $\prescript{\mathrm{lo}}{}{\sigma}_\gamma$ & \SI{0.001}{rad} \\
    $\prescript{\mathrm{hi}}{}{\sigma}_\gamma$ & \SI{1.0}{rad} \\

    $\prescript{\mathrm{Vmin}}{}{\sigma}_\mathrm{3D}$ & \SI{0.1}{m} \\
    $\prescript{\mathrm{Vmax}}{}{\sigma}_\mathrm{3D}$ & \SI{5.0}{m} \\
    $\nu$ & \num{400} \\

    $\mu_Y$ & \num{260} \\
    $\mu_Z$ & \num{500} \\
    $\prescript{\mathrm{V}}{}{\sigma}_\gamma$ & \SI{0.01}{rad} \\
    \bottomrule
\end{tabularx}
  \label{tab:params}
\end{center}
\end{table}

\noindent The proposed cooperative localization approach was quantitatively evaluated on custom datasets gathered in an indoor environment with motion-capture ground-truth data and in a large-scale outdoor environment with \ac{RTK} ground-truth data.

To the best of the authors' knowledge, there are no existing, openly-available algorithms utilizing the same hardware setup, and alternative cooperative localization approaches require fundamentally different hardware on either one or all robots, making direct experimental comparison infeasible.
Therefore, the following quantitative evaluation of the proposed approach focuses mainly on the localization improvements provided by the cooperative localization method with respect to the individual odometry algorithms on the specific robots.

In all evaluations, the LIO-SAM algorithm~\cite{shanLIOSAMTightlycoupledLidar2020} was used as the \ac{LIO} of robot $X$.
The LIO-SAM algorithm was modified to work with the six-axis \ac{IMU} of the Ouster 3D \ac{lidar}; thus, the \ac{LIO} does not have information about the global yaw orientation.
The internal threshold of LIO-SAM for considering the scan matching problem degenerated was empirically tuned for use with the employed 3D \ac{lidar}.
The OpenVINS algorithm~\cite{genevaOpenVINSResearchPlatform2020a} was utilized as the \ac{VIO} of robot $Y$.
Parameters of OpenVINS were tuned for reliable performance on the real-world data.
Note that the parameters of OpenVINS were not the same on all datasets due to the use of different cameras in the indoor and outdoor experiments, different constructions of the camera-equipped \acp{UAV}, and different amounts of visual features in the environments.
For obtaining the relative detections of robot positions, we placed reflective markers on the \acp{UAV} representing the detected robots $Y$ and $Z$ and we utilized a 3D \ac{lidar}-based detector inspired by the tracking module of~\cite{vrbaOnboardLiDARBasedFlying2025} with the detections initialized based on the reflective markers and using centroids of clusters of \ac{lidar} points as the robot detections.
The system time of the robots was synchronized over a wireless network using the chrony implementation of the \ac{NTP}.
The sensor data were recorded on board the robots and processed offline.
Table~\ref{tab:params} shows data rates and parameters used by the proposed algorithm.
These parameter values were the same for all experiments.

  When processing the ground-truth measurements and evaluating the accuracy of a robot trajectory, we followed the approach proposed in~\cite{zhangTutorialQuantitativeTrajectory2018}.
For each individual estimated trajectory, we calculated ground-truth data interpolated to the times of the estimate, aligned the estimated data to the ground-truth data using the Kabsch-Umeyama algorithm~\cite{umeyamaLeastsquaresEstimationTransformation1991}, and calculated the \ac{ATE} as the \ac{RMSE} of the estimated position/yaw orientation.
In all plots of the estimated data, the estimates were plotted in the ground-truth reference frame after the alignment.

\subsection{Real-World UGV-UAV Indoor Experiments}
  \label{sec:exp_turku}

\begin{table}[t]
\scriptsize
\begin{center}
  \caption{Positional and rotational localization errors from the experimental evaluation. Localization errors of the proposed cooperative localization method (COOP) and of the respective individual odometry methods (LIO, VIO) are compared. The minimal errors for each robot in each dataset are in bold.}
  \begin{tabularx}{1.0\linewidth}{p{1.2cm} X X X X X} 
\toprule
\toprule
  \textbf{Dataset} & \textbf{Robot} & \textbf{Method}  & \textbf{2D ATE [m]} & \textbf{3D ATE [m]} & \textbf{rot. ATE [rad]} \\
\midrule
  \multirow[c]{4}{*}{\makecell{Indoor \#1\\circle\\var. height}} & \multirow[c]{2}{*}{UGV $X$} & LIO & \bfseries 0.029 & \bfseries 0.047 & \bfseries 0.013 \\
 &  & COOP & \bfseries 0.029 & \bfseries 0.047 & \bfseries 0.013 \\
  \cmidrule(lr){2-6}
 & \multirow[c]{2}{*}{UAV $Y$} & VIO & 0.483 & 0.489 & 0.155 \\
 &  & COOP & \bfseries 0.065 & \bfseries 0.079 & \bfseries 0.041 \\

  \cmidrule(lr){1-6}
  \multirow[c]{4}{*}{\makecell{Indoor \#2\\circle\\var. height\\VIO deg.}} & \multirow[c]{2}{*}{UGV $X$} & LIO & \bfseries 0.029 & \bfseries 0.047 & \bfseries 0.013 \\
 &  & COOP & 0.030 & 0.068 & \bfseries 0.013 \\
  \cmidrule(lr){2-6}
 & \multirow[c]{2}{*}{UAV $Y$} & VIO & 0.497 & 4.671 & 0.104 \\
 &  & COOP & \bfseries 0.076 & \bfseries 0.101 & \bfseries 0.043 \\
 
\cmidrule(lr){1-6}
  \multirow[c]{4}{*}{\makecell{Indoor \#3\\figure eight}} & \multirow[c]{2}{*}{UGV $X$} & LIO & \bfseries 0.025 & \bfseries 0.029 & \bfseries 0.010 \\
 &  & COOP & 0.026 & \bfseries 0.029 & \bfseries 0.010 \\
  \cmidrule(lr){2-6}
 & \multirow[c]{2}{*}{UAV $Y$} & VIO & 0.280 & 0.281 & \bfseries 0.012 \\
 &  & COOP & \bfseries 0.076 & \bfseries 0.089 & 0.013 \\

\cmidrule(lr){1-6}
  \multirow[c]{4}{*}{\makecell{Indoor \#4\\circle}} & \multirow[c]{2}{*}{UGV $X$} & LIO & \bfseries 0.023 & \bfseries 0.044 & \bfseries 0.017 \\
 &  & COOP & 0.027 & 0.046 & \bfseries 0.017 \\
  \cmidrule(lr){2-6}
 & \multirow[c]{2}{*}{UAV $Y$} & VIO & 0.440 & 0.458 & 0.137 \\
 &  & COOP & \bfseries 0.084 & \bfseries 0.098 & \bfseries 0.048 \\

 \midrule
 \midrule
  \multirow[c]{4}{*}{\makecell{Outdoor \#1\\around\\field}} & \multirow[c]{2}{*}{UAV $X$} & LIO & 74.386 & 74.591 & 1.219 \\
 &  & COOP & \bfseries 10.091 & \bfseries 10.373 & \bfseries 0.360 \\
  \cmidrule(lr){2-6}
 & \multirow[c]{2}{*}{UAV $Y$} & VIO & 46.391 & 46.458 & 0.131 \\
 &  & COOP & \bfseries 10.425 & \bfseries 10.692 & \bfseries 0.092 \\

\cmidrule(lr){1-6}
  \multirow[c]{4}{*}{\makecell{Outdoor \#2\\between\\houses}} & \multirow[c]{2}{*}{UAV $X$} & LIO & \bfseries 0.145 & 0.972 & \bfseries 0.154 \\
 &  & COOP & 0.155 & \bfseries 0.902 & 0.156 \\
  \cmidrule(lr){2-6}
 & \multirow[c]{2}{*}{UAV $Y$} & VIO & 0.553 & \bfseries 0.715 & 0.112 \\
 &  & COOP & \bfseries 0.162 & 0.896 & \bfseries 0.109 \\

\cmidrule(lr){1-6}
  \multirow[c]{6}{*}{\makecell{Outdoor \#3\\3 UAVs\\}} & \multirow[c]{2}{*}{UAV $X$} & LIO & \bfseries 0.194 & \bfseries 0.200 & 0.052 \\
 &  & COOP & 0.198 & 0.204 & \bfseries 0.051 \\
  \cmidrule(lr){2-6}
 & \multirow[c]{2}{*}{UAV $Y$} & VIO & \bfseries 0.295 & 0.923 & 0.069 \\
 &  & COOP & 0.315 & \bfseries 0.333 & \bfseries 0.063 \\
  \cmidrule(lr){2-6}
 & \multirow[c]{2}{*}{UAV $Z$} & VIO & 0.595 & 1.010 & \bfseries 0.197 \\
 &  & COOP & \bfseries 0.311 & \bfseries 0.328 & 0.239 \\

\cmidrule(lr){1-6}
  \multirow[c]{6}{*}{\makecell{Outdoor \#4\\3 UAVs\\LIO deg.}} & \multirow[c]{2}{*}{UAV $X$} & LIO & 4.076 & 4.077 & 1.668 \\
 &  & COOP & \bfseries 0.525 & \bfseries 0.875 & \bfseries 0.067 \\
  \cmidrule(lr){2-6}
 & \multirow[c]{2}{*}{UAV $Y$} & VIO & \bfseries 0.294 & 0.923 & 0.069 \\
 &  & COOP & 0.354 & \bfseries 0.767 & \bfseries 0.060 \\
  \cmidrule(lr){2-6}
 & \multirow[c]{2}{*}{UAV $Z$} & VIO & 0.595 & 1.010 & \bfseries 0.197 \\
 &  & COOP & \bfseries 0.550 & \bfseries 0.880 & 0.210 \\
\bottomrule

\end{tabularx}
  \label{tab:rmse}
\end{center}
\end{table}

\begin{figure*}[t]
  \centering
  \begin{tikzpicture}
    \node[anchor=north west,inner sep=0] (a) at (0, 0)
    {
      \includegraphics[width=0.325\linewidth, trim=0cm 0cm 0cm 0cm, clip=true]{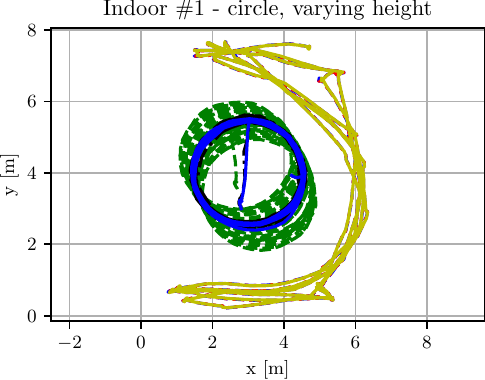}
    };

    \node[anchor=north west,inner sep=0] (b) at (5.8cm, 0cm)
    {
      \includegraphics[width=0.325\linewidth, trim=0cm 0cm 0cm 0cm, clip=true]{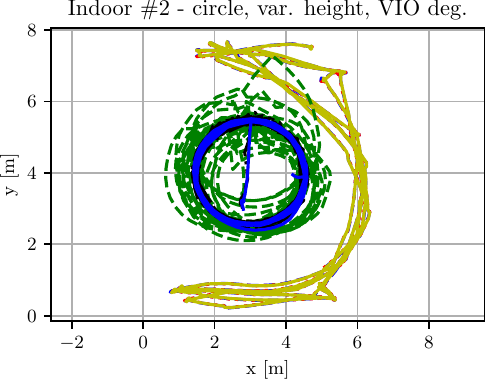}
    };

    \node[anchor=north west,inner sep=0] (c) at (11.6cm, 0cm)
    {
      \includegraphics[width=0.325\linewidth, trim=0cm 0cm 0cm 0cm, clip=true]{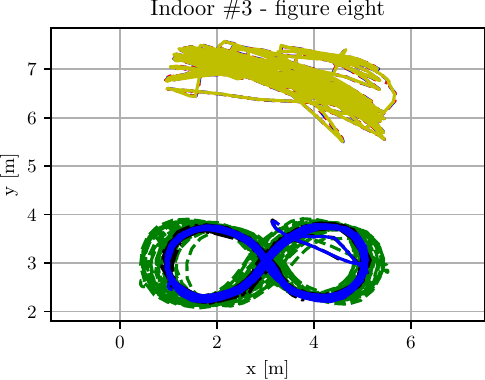}
    };

    \node[anchor=north west,inner sep=0] (d) at (0, -4.7cm)
    {
      \includegraphics[width=0.325\linewidth, trim=0cm 0cm 0cm 0cm, clip=true]{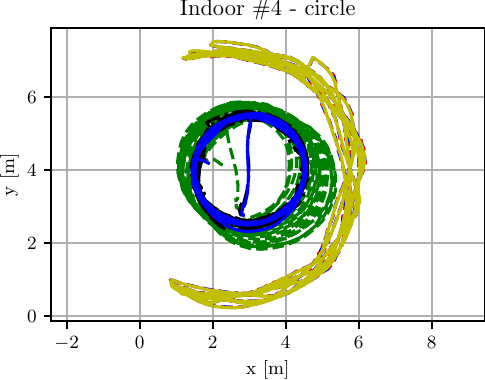}
    };

    \node[anchor=north west,inner sep=0] (e) at (5.8cm, -4.7cm)
    {
      \includegraphics[width=0.325\linewidth, trim=0cm 0cm 0cm 0cm, clip=true]{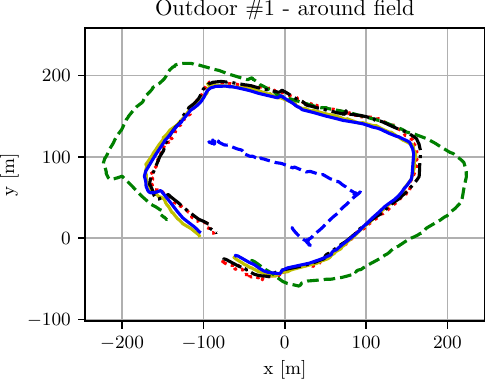}
    };

    \node[anchor=north west,inner sep=0] (f) at (11.6cm, -4.7cm)
    {
      \includegraphics[width=0.325\linewidth, trim=0cm 0cm 0cm 0cm, clip=true]{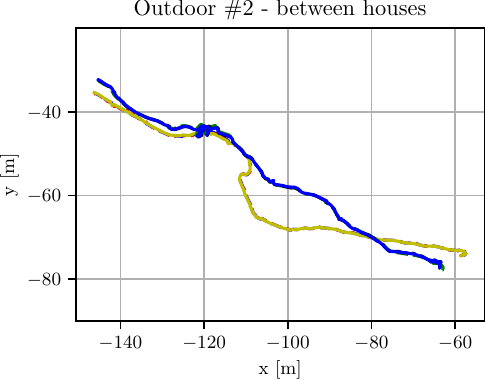}
    };

    \node[anchor=north west,inner sep=0] (g) at (0, -9.4cm)
    {
      \includegraphics[width=0.325\linewidth, trim=0cm 0cm 0cm 0cm, clip=true]{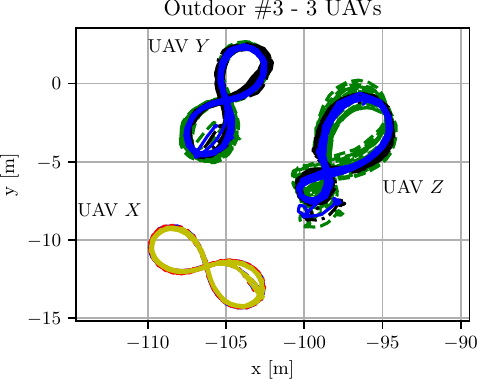}
    };

    \node[anchor=north west,inner sep=0] (h) at (5.8cm, -9.4cm)
    {
      \includegraphics[width=0.325\linewidth, trim=0cm 0cm 0cm 0cm, clip=true]{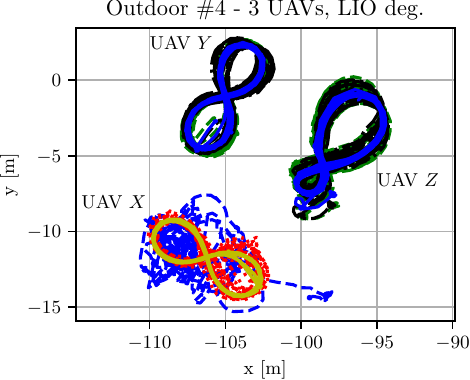}
    };

    \node[anchor=north west,inner sep=0] (i) at (13.0cm, -9.4cm)
    {
      \includegraphics[width=0.21\linewidth, trim=3.9cm 0cm 0cm 1cm, clip=true]{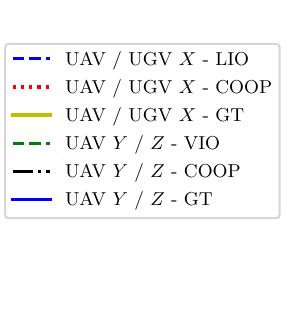}
    };

  \end{tikzpicture}
  \caption{$xy$-plots of odometry data (LIO, VIO), the proposed cooperative localization method (COOP), and ground truth (GT) from the experimental evaluation. All estimated trajectories were aligned to the ground-truth data for evaluation.}
  \label{fig:exp_plots_xy}
\end{figure*}

\noindent As the \ac{lidar}-carrying robot $X$, we utilized the Unitree B1 quadruped carrying the Ouster OS0-128 \ac{lidar}.
As the camera-equipped robot $Y$, we used a small custom-made \ac{UAV} based on the DJI F330 frame, equipped with the RealSense T265 tracking camera and with reflective markers placed on its legs (see Fig.~\ref{fig:motivation}).
Ground-truth measurements were obtained using the OptiTrack motion capture system.

We performed four different experiments.
  The localization errors from these experiments are listed in Table~\ref{tab:rmse}.
  The $xy$-plots of data obtained in these experiments are shown in Fig.~\ref{fig:exp_plots_xy}, and the altitude and yaw plots are shown in Fig.~\ref{fig:exp_plots_zyaw}.
  The experiments were performed in a feature-rich environment, where the \ac{LIO} was considered reliable the entire time.

  In the \textit{Indoor \#1} experiment, the \ac{UAV} $Y$ was flying in a circle with varying altitude, while the \ac{UGV} $X$ was walking around the circle.
  For the \ac{UGV}, the error of the proposed cooperative localization method (COOP) was the same as the error of the \ac{LIO}, as the \ac{LIO} was considered reliable the entire time.
  For the \ac{UAV}, both the positional and rotational errors were significantly reduced.

  In the \textit{Indoor \#2} experiment, we utilized the same data as in the previous case, but we artificially created strong \ac{VIO} degradation in part of the trajectory.
  For half of the \ac{UAV}'s trajectory (when $x > 3$), the camera images were replaced with completely black images.
  In the degraded part of the trajectory, the \ac{VIO} needed to rely on \ac{IMU} data only and therefore exhibited significant drift, mainly in the $z$-axis.
  For the \ac{UGV}, the COOP error stayed approximately the same as the \ac{LIO} error.
  For the \ac{UAV}, the VIO error was corrected by the COOP method, with the 3D \ac{ATE} decreasing from \SI{4.671}{m} to \SI{0.101}{m}.

  In the \textit{Indoor \#3} and \textit{Indoor \#4} experiments, the \ac{UAV} $X$ was flying in a figure-eight and circular trajectory, respectively, with the \ac{UGV} walking next to it.
  For the \ac{UGV}, the COOP error again stayed approximately the same as the error of \ac{LIO}.
  For the \ac{UAV}, the positional error was significantly reduced.
  In the figure-eight trajectory, the \ac{UAV}'s  VIO rotational error was already very low, and the COOP rotational error stayed approximately the same.
  In the circular trajectory, the \ac{UAV}'s rotational error was significantly reduced.

\begin{figure*}[t]
  \centering
  \begin{tikzpicture}
    \node[anchor=north west,inner sep=0] (a) at (0, 0)
    {
      \includegraphics[width=0.325\linewidth, trim=0cm 0cm 0cm 0cm, clip=true]{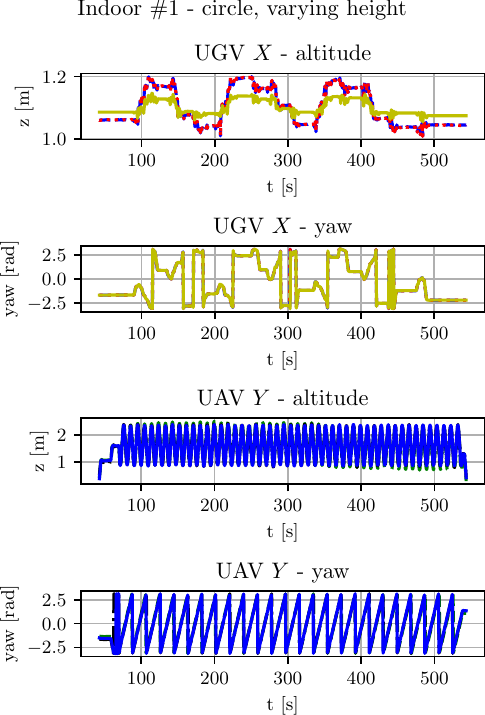}
    };

    \node[anchor=north west,inner sep=0] (b) at (5.8cm, 0cm)
    {
      \includegraphics[width=0.325\linewidth, trim=0cm 0cm 0cm 0cm, clip=true]{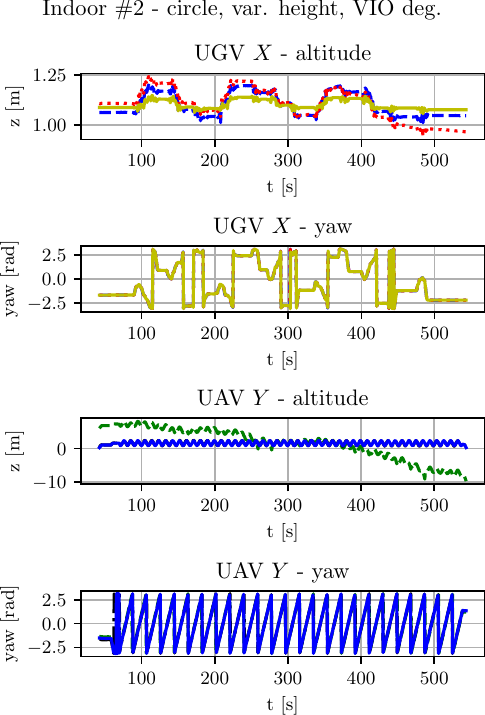}
    };

    \node[anchor=north west,inner sep=0] (c) at (11.6cm, 0cm)
    {
      \includegraphics[width=0.325\linewidth, trim=0cm 0cm 0cm 0cm, clip=true]{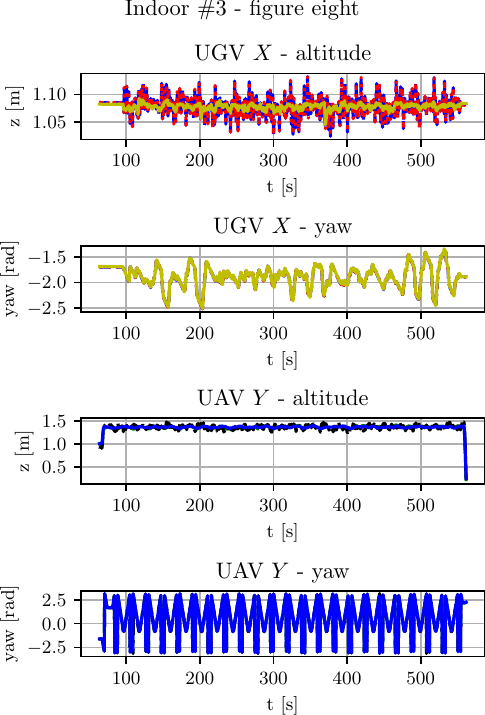}
    };

    \node[anchor=north west,inner sep=0] (d) at (0, -9.3cm)
    {
      \includegraphics[width=0.325\linewidth, trim=0cm 0cm 0cm 0cm, clip=true]{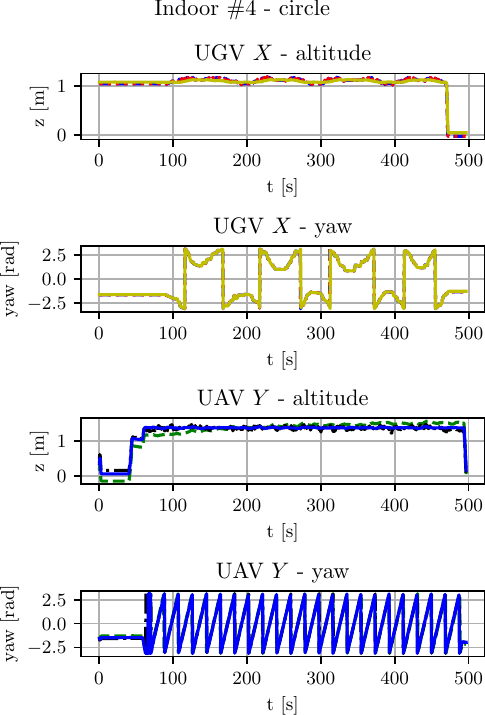}
    };

    \node[anchor=north west,inner sep=0] (e) at (5.8cm, -9.3cm)
    {
      \includegraphics[width=0.325\linewidth, trim=0cm 0cm 0cm 0cm, clip=true]{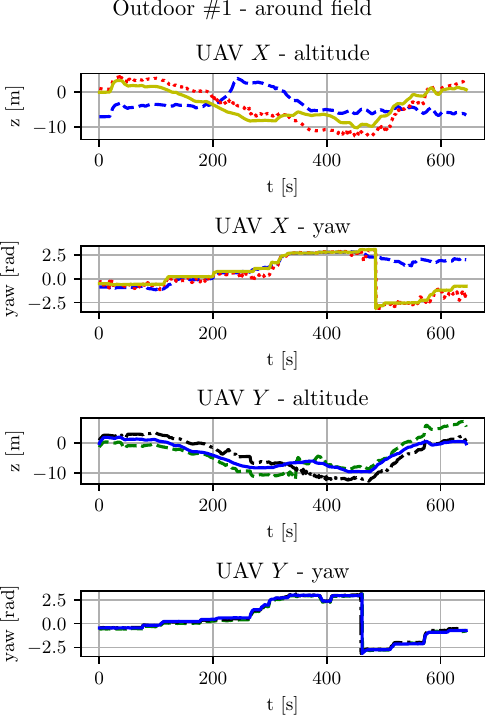}
    };

    \node[anchor=north west,inner sep=0] (f) at (11.6cm, -9.3cm)
    {
      \includegraphics[width=0.325\linewidth, trim=0cm 0cm 0cm 0cm, clip=true]{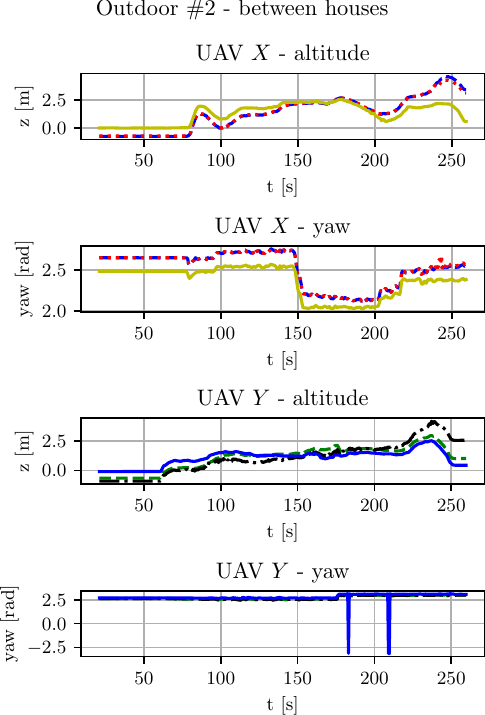}
    };

    \node[anchor=north west,inner sep=0] (c) at (0cm, -18cm)
    {
      \includegraphics[width=1.0\linewidth, trim=0cm 11.3cm 0cm 0cm, clip=true]{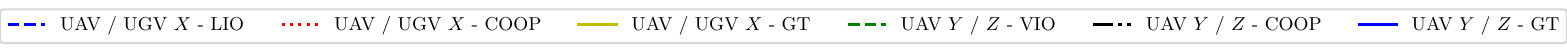}
    };

  \end{tikzpicture}
  \caption{Altitude and yaw plots of odometry data (LIO, VIO), the proposed cooperative localization method (COOP), and ground-truth data (GT).}
  \label{fig:exp_plots_zyaw}
\end{figure*}

\begin{figure*}[ht]
  \centering
  \begin{tikzpicture}
    \node[anchor=north west,inner sep=0] (a) at (0, 0)
    {
      \includegraphics[width=0.325\linewidth, trim=0cm 0cm 0cm 0cm, clip=true]{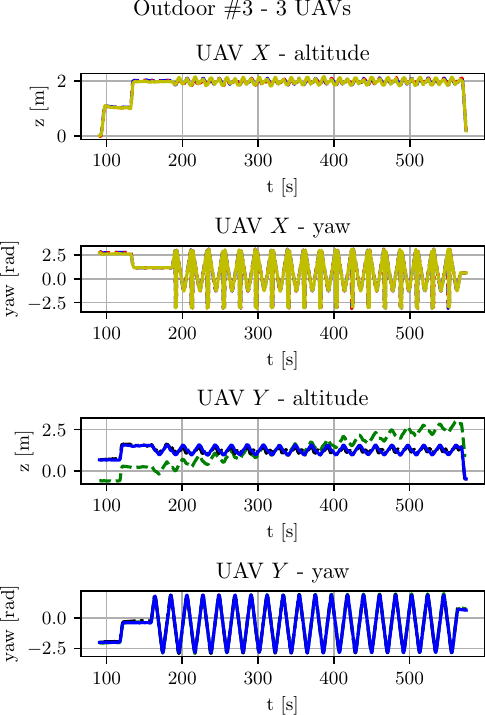}
    };

    \node[anchor=north west,inner sep=0] (b) at (5.8cm, 0cm)
    {
      \includegraphics[width=0.325\linewidth, trim=0cm 0cm 0cm 0cm, clip=true]{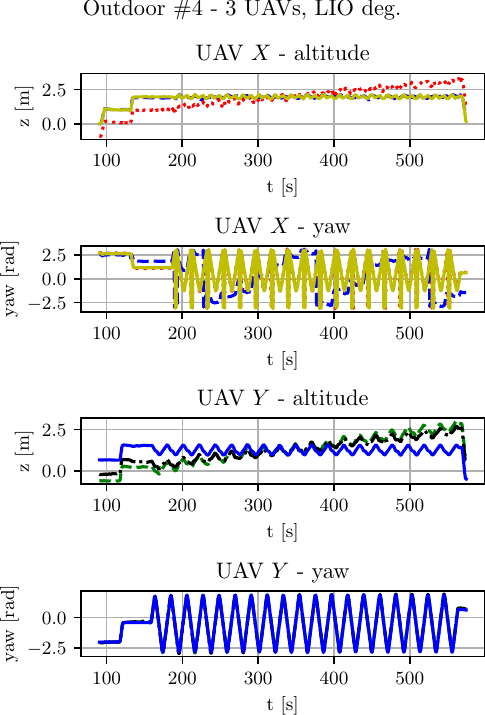}
    };

    \node[anchor=north west,inner sep=0] (c) at (11.6cm, 0cm)
    {
      \includegraphics[width=0.325\linewidth, trim=0cm 0cm 0cm 0cm, clip=true]{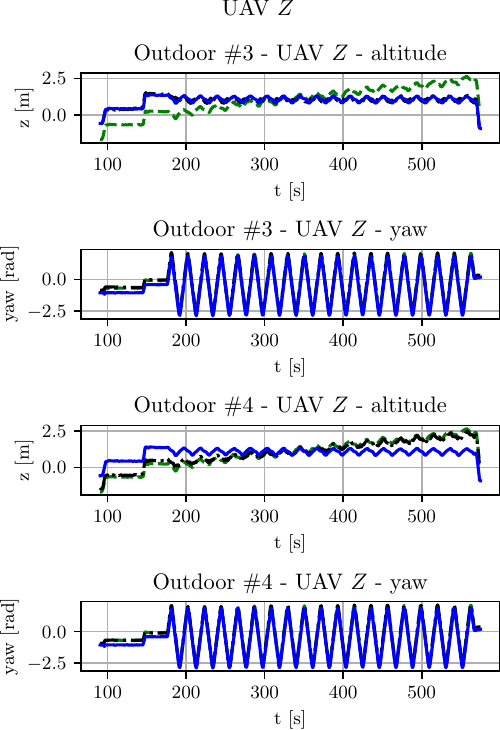}
    };

    \node[anchor=north west,inner sep=0] (c) at (0cm, -8.7cm)
    {
      \includegraphics[width=1.0\linewidth, trim=0cm 11.3cm 0cm 0cm, clip=true]{figures/exp_zyaw8.pdf}
    };

  \end{tikzpicture}
  \caption{Continuation of the altitude and yaw plots of odometry data (LIO, VIO), the proposed cooperative localization method (COOP), and ground truth data (GT).}
  \label{fig:exp_plots_zyaw2}
\end{figure*}

\subsection{Real-World Multi-UAV Outdoor Experiments}
\label{sec:exp_temesvar}

\noindent A custom-made \ac{UAV} based on the X500 frame was used as the robot $X$~\cite{hertMRSDroneModular2023a}.
The \ac{UAV} carried the Ouster OS0-128 \ac{lidar}.
As the camera-equipped robots $Y$ and $Z$, we utilized two smaller \acp{UAV} equipped with the mvBlueFOX-MLC200wG cameras and the ICM-42688-P \ac{IMU}, providing data for the \ac{VIO} (see Fig.~\ref{fig:motivation}).
The smaller \acp{UAV} each had a slightly different construction, resulting in different amounts of propeller-induced vibrations acting upon the \ac{IMU}, thus resulting in different \ac{VIO} accuracy for each \ac{UAV}.
Ground-truth positional measurements were provided by the Holybro H-RTK F9P modules.
Ground-truth measurements of yaw were obtained from the onboard flight controller units processing external magnetometer data from the F9P modules.
The localization errors from the outdoor experiments are listed in Table~\ref{tab:rmse}, the $xy$-plots are shown in Fig.~\ref{fig:exp_plots_xy}, and the altitude and yaw plots are shown in Fig.~\ref{fig:exp_plots_zyaw} and Fig.~\ref{fig:exp_plots_zyaw2}.

\begin{figure}[t]
  \centering
    \input{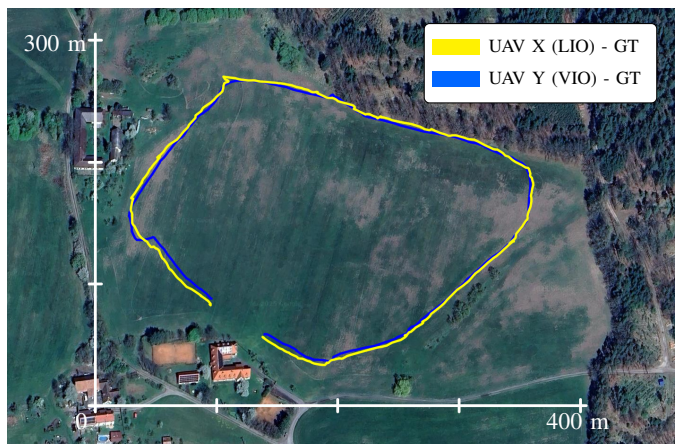}
  \caption{Ground-truth trajectories from the \textit{Outdoor \#1} experiment, where the two \acp{UAV} flew around an open field.}
  \label{fig:around_field_gt}
\end{figure}

Four different experiments were made.
In the \textit{Outdoor \#1}, the \acp{UAV} $X$ and $Y$ flew around a large-scale open field (see Fig.~\ref{fig:around_field_gt}).
The mutual distance between the \acp{UAV} ranged between \SI{2.9}{m} and \SI{24.9}{m} with the mean mutual distance of \SI{9.9}{m}.
The experiment was characterized by large portions of \ac{LIO} degeneracy.
The \ac{LIO} was reliable only for limited times, mainly at the right and top parts of the trajectory, when the \acp{UAV} flew close to the trees next to the field.
The rest of the time, the \ac{lidar} data contained only the featureless ground plane with no geometric features usable to sufficiently constrain the scan matching problem.
Although the \ac{VIO} of \ac{UAV} $Y$ was usable the entire time, it exhibited significant long-term drift, significantly increasing the scale of the entire \ac{VIO} trajectory (see Fig.~\ref{fig:exp_plots_xy} - Outdoor \#1).
The proposed cooperative localization method managed to combine the data, significantly decreasing the localization errors for both \acp{UAV}.
The localization error of \ac{UAV} $X$ decreased from \SI{74.591}{m} to \SI{10.373}{m} in position and from \SI{1.219}{rad} to \SI{0.360}{rad} in yaw.
The error of \ac{UAV} $Y$ decreased from \SI{46.458}{m} to \SI{10.692}{m} in 3D position and from \SI{0.131}{rad} to \SI{0.092}{rad} in yaw.

In the \textit{Outdoor \#2} experiment, the \acp{UAV} flew in an environment between several buildings, where the \ac{LIO} was reliable for the entire time.
The COOP error of \ac{UAV} $X$ stayed approximately the same as the \ac{LIO} error.
The COOP error of \ac{UAV} $Y$ significantly decreased in 2D, but slightly increased in 3D due to the altitude estimation error of the \ac{LIO} at the end of the trajectory.

In the \textit{Outdoor \#3} experiment, one \ac{lidar}-equipped \ac{UAV} $X$ and two camera-equipped \acp{UAV} $Y$ and $Z$ were utilized.
The \acp{UAV} were flying in figure-eight trajectories, and the \ac{LIO} of \ac{UAV} $X$ was reliable the entire time.
For the \ac{UAV} $X$, the localization error of COOP stayed approximately the same as the \ac{LIO} localization error.
For the \ac{UAV} $Y$, the error of COOP was similar to the \ac{VIO} error with significant error decrease in the $z$-axis.
For the \ac{UAV} $Z$, there was a significant decrease in both the 2D and 3D error.

The \textit{Outdoor \#4} experiment utilized the same raw data as the previous experiment.
In this experiment, we artificially created \ac{LIO} degradation.
After the \acp{UAV} started following their trajectories, the point clouds incoming into the \ac{LIO} algorithm were modified by removing all points above the ground plane, thus removing features usable for scan matching.
This feature removal resulted in the \ac{LIO} data being useless in the $xy$-plane (see Fig.~\ref{fig:exp_plots_xy} - Outdoor \#4), as well as in yaw (see Fig.~\ref{fig:exp_plots_zyaw2} - Outdoor \#4).
Utilizing the \ac{VIO} data from the remaining two \acp{UAV} and their detections, the proposed cooperative localization method was able to correctly estimate the pose of \ac{UAV} $X$ and reconstruct its figure-eight trajectory.
The localization error of \ac{UAV} $X$ was reduced from \SI{4.077}{m} to \SI{0.875}{m} in 3D position and from \SI{1.668}{rad} to \SI{0.067}{rad} in yaw orientation.
It is worth mentioning that the use of a single detected \ac{UAV} was insufficient for correctly estimating the changing pose of \ac{UAV} $X$ in the presence of such strong \ac{LIO} degradation.
This corresponds to the findings of the observability analysis in sec.~\ref{sec:obs_lio}, which showed that in the case of a \ac{LIO} degeneracy and a single detected robot, there is one unobservable \ac{DOF}, while in the case of two detected robots, this unobservability is mitigated.

\subsection{Evaluating the Use of Wasserstein Distance}
\label{sec:exp_wasserstein}

\subsubsection{Correlation Between the Wasserstein Distance and Localization Error}

\noindent We want to verify the assumption that the Wasserstein distance between the covariance matrices of two consecutive \ac{VIO} positions correlates with the real-world localization error and validate the decision to set the standard deviations of the \ac{VIO} relative position measurements proportional to the Wasserstein distance.
For each experiment and each respective camera-equipped \ac{UAV}, we calculated the Pearson correlation coefficient of the Wasserstein distance corresponding to each relative \ac{VIO} position measurement and the norm of the 3D relative position error, calculated from the ground-truth data.
Table~\ref{tab:wasser_corr} shows the obtained correlation coefficients.
A positive correlation between the Wasserstein distance and relative position error was found in all datasets.
The largest correlations were observed in the \textit{Indoor \#2} and \textit{Outdoor \#1} experiments, which contained the largest variations in \ac{VIO} accuracy.
In the \textit{Indoor \#2} experiment, we repeatedly created \ac{VIO} degradations by blocking the camera image, and in the \textit{Outdoor \#1} experiment, the number of visual features varied significantly while the \ac{UAV} was flying around the field.

\begin{table}[t]
\scriptsize
\begin{center}
  \caption{Pearson correlation coefficients of Wasserstein distances and the norms of relative 3D position error of the \ac{VIO}. The values for Outdoor \#4 are equal to the values of Outdoor \#3, as the VIO data were equal.}
  \begin{tabularx}{1.0\linewidth}{X X} 
    \toprule 
    \textbf{Dataset} & \textbf{Correlation coefficient} \\ \midrule

    Indoor \#1 & \num{0.221} \\
    Indoor \#2 & \num{0.758} \\
    Indoor \#3 & \num{0.590} \\
    Indoor \#4 & \num{0.563} \\
    Outdoor \#1 & \num{0.807} \\
    Outdoor \#2 & \num{0.644} \\
    Outdoor \#3 - UAV $X$ & \num{0.460} \\
    Outdoor \#3 - UAV $Y$ & \num{0.597} \\
    \bottomrule
\end{tabularx}
  \label{tab:wasser_corr}
\end{center}
\end{table}

\subsubsection{Performance in the Outdoor \#1 Dataset}

\begin{figure}[t]
  \centering

  \begin{tikzpicture}
    \node[anchor=north west,inner sep=0] (a) at (0, 0)
    {
      \includegraphics[width=1.0\linewidth, trim=0cm 0cm 0cm 0cm, clip=true]{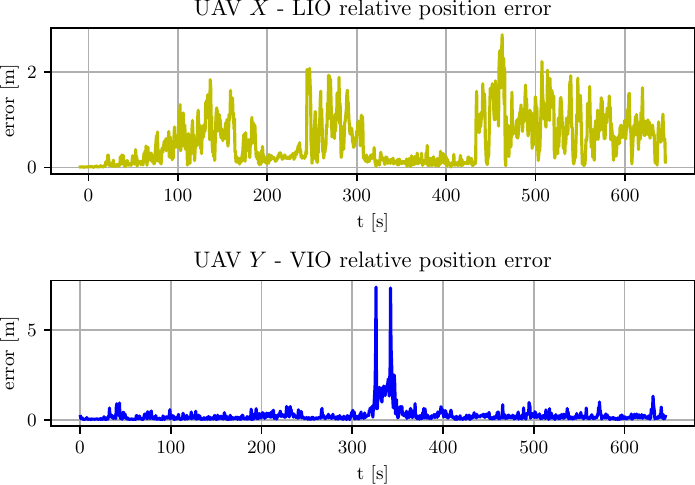}
    };

    \node[anchor=north west,inner sep=0] (a) at (0, -6.3)
    {
      \includegraphics[width=1.0\linewidth, trim=0cm 0cm 0cm 0cm, clip=true]{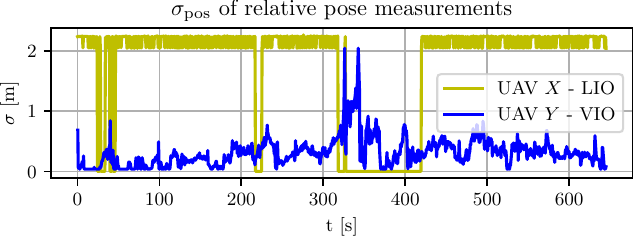}
    };

  \end{tikzpicture}

  \caption{Comparison of the \ac{LIO} relative position error, \ac{VIO} relative position error, and the $\sigma_\mathrm{pos}$ values used in the covariance matrices of the proposed cooperative localization method from the \textit{Outdoor \#1} experiment.}
  \label{fig:exp_rel_errors}
\end{figure}

  Fig.~\ref{fig:exp_rel_errors} shows a comparison of the \ac{VIO} and \ac{LIO} relative position errors and the $\sigma_\mathrm{pos}$ utilized by the proposed cooperative localization method from the \textit{Outdoor \#1} experiment.
  This data showcases the necessity of having the measurement covariance matrices adaptively react to changing odometry accuracy.
  The \ac{LIO} relative position error plot shows several moments in time when the \ac{LIO} error was low, surrounded by periods of low accuracy.
  The $\sigma_\mathrm{pos}$ of \ac{UAV} $X$ in the bottom plot reflects this changing accuracy.

  Similarly, the $\sigma_\mathrm{pos}$ of \ac{UAV} $Y$ reflects the changing localization accuracy of the \ac{VIO}.
  Notably, there was a large increase of \ac{VIO} localization error after $t = \SI{320}{s}$ with two sharp peaks at $t=\SI{326}{s}$ and $t=\SI{342}{s}$.
  Failing to detect such an increase in localization error would lead to inconsistency between the measurements obtained from the detections and the \ac{VIO} data and would result in large estimation errors.
  In our experiments, utilizing a static covariance matrix, which did not reflect this large localization error, resulted in loss of the estimate due to large localization error and subsequent failure to associate new detections to the estimate due to exceeding the maximum allowed detection distance.
  It is worth mentioning that such sharp peaks in localization error and in the Wasserstein distance are often produced after loss and subsequent re-detection of visual features.
  The employed \ac{VIO} algorithm, OpenVINS, tracks a fixed number of \ac{SLAM} features in its state vector throughout the estimation process.
  After the features are lost, the correction caused by regaining the features can cause a sudden jump in the \ac{VIO} estimate, accompanied by a sudden decrease in the covariance matrix of the \ac{VIO} position.
  The Wasserstein distance can be used to conveniently quantify such a change in localization accuracy and lower the weight of the corresponding relative position measurement in the factor graph.

  \subsubsection{Ablation Study on Wasserstein Distance}

  To verify the importance of the Wasserstein distance use, we performed an ablation study on the \textit{Outdoor \#1} dataset, comparing the use of adaptive $\prescript{\mathrm{V}}{}{\sigma}^2_\mathrm{pos}$ calculated using eq.~(\ref{eq:vio_sigma}) and the use of static variance adapting only to the sampling rate, calculated as
  \begin{equation}
    \prescript{\mathrm{V}}{}{\sigma}^2_\mathrm{pos} = \frac{\Delta t}{3} \prescript{\mathrm{V}}{}{\sigma}^2_\mathrm{3D}.
  \end{equation}
Furthermore, the static variant had the yaw variance inflation from eq.~(\ref{eq:vio_yaw_inflation}) disabled.
We ran the adaptive variant of the algorithm for 100 values of $\mu$, logarithmically-spaced between \num{1.0} and \num{1.0e4}, and the static algorithm for 100 values of $\prescript{\mathrm{V}}{}{\sigma}_\mathrm{3D}$, logarithmically-spaced between \num{0.1} and \SI{1.0}{m}.

\begin{figure}[t]
  \centering
  \includegraphics[width=1.0\linewidth, trim=0.0cm 0.0cm 0.0cm 0.0cm, clip=true]{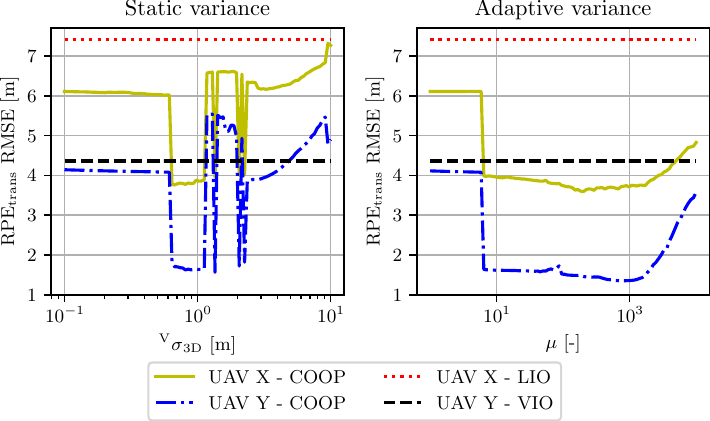}
  \caption{Comparison of the RMSE of the translational RPE at the traveled distance of \SI{10}{m} for static and adaptive setting of the \ac{VIO} variance for the entire evaluated parameter ranges.
  }

  \label{fig:wasserstein_rpe_plot}
\end{figure}

The main advantage of using the Wasserstein distance in this case lies in the ability to react to local changes of \ac{VIO} accuracy.
With static variance, the estimated \ac{UAV} positions exhibited significant oscillations due to the decrease in \ac{VIO} accuracy after $t=\SI{320}{s}$ and subsequent discrepancy between the weights of the \ac{LIO} and \ac{VIO} data.
To quantitatively evaluate this difference in local accuracy, we have calculated the \ac{RPE}~\cite{zhangTutorialQuantitativeTrajectory2018} of the method at different traveled distances and compared the \ac{RMSE} of the translational part of the \ac{RPE} among the different approaches.
Fig.~\ref{fig:wasserstein_rpe_plot} shows the dependency of the \ac{RMSE} of the translational part of the \ac{RPE} at the traveled distance of \SI{10}{m} on the parameter values for both the static and adaptive variance setting.
With the use of adaptive variance setting based on the Wasserstein distance, the algorithm was able to achieve lower \ac{RPE} and exhibited more stable performance compared to the static variant.
For many values of the static variance, the algorithm lost track of the detected \ac{UAV} due to the large discrepancy between the estimated poses of the \acp{UAV} and subsequently exceeding the detection association threshold.
Table~\ref{tab:wasser_rpe} shows the specific values of the minimum achievable translational \ac{RPE} at the different traveled distances.
The adaptive method, utilizing the Wasserstein distance, outperformed the use of static variance at all of the evaluated distances.

\begin{table}[t]
\small
\scriptsize
\begin{center}
  \caption{Comparison of minimum achievable Relative Pose Error (RPE) of both \acp{UAV} for each method at different traveled distances.}
  \begin{tabularx}{1.0\linewidth}{p{0.7cm} p{1.0cm} X X X X X} 
    \toprule 
    \multicolumn{2}{c}{\textbf{Distance traveled [m]}}  & \textbf{5} & \textbf{10} & \textbf{20} & \textbf{50} & \textbf{100} \\ \midrule
    \textbf{Method} & \textbf{Robot}  & \multicolumn{5}{c}{\textbf{minimal RMSE of translational RPE [m]}} \\ \midrule
    \multirow[c]{2}{*}{Static} & UAV $X$ & 2.48 & 3.76 & 6.72 & 16.33 & 30.66 \\
    & UAV $Y$ & 1.11 & 1.57 & 2.44 & 4.99 & 8.81 \\
    \multirow[c]{2}{*}{Adaptive} & UAV $X$ & \textbf{2.38} & \textbf{3.59} & \textbf{6.44} & \textbf{15.49} & \textbf{28.81} \\
    & UAV $Y$ & \textbf{0.90} & \textbf{1.35} & \textbf{2.18} & \textbf{4.53} & \textbf{8.10} \\

    \bottomrule
\end{tabularx}
  \label{tab:wasser_rpe}
\end{center}
\end{table}

\subsection{Investigating the Findings of the Observability Analysis}
\label{sec:exp_observability}

\begin{figure*}[t]
  \centering
  \includegraphics[width=1.0\linewidth, trim=0.0cm 0.0cm 0.0cm 0.0cm, clip=true]{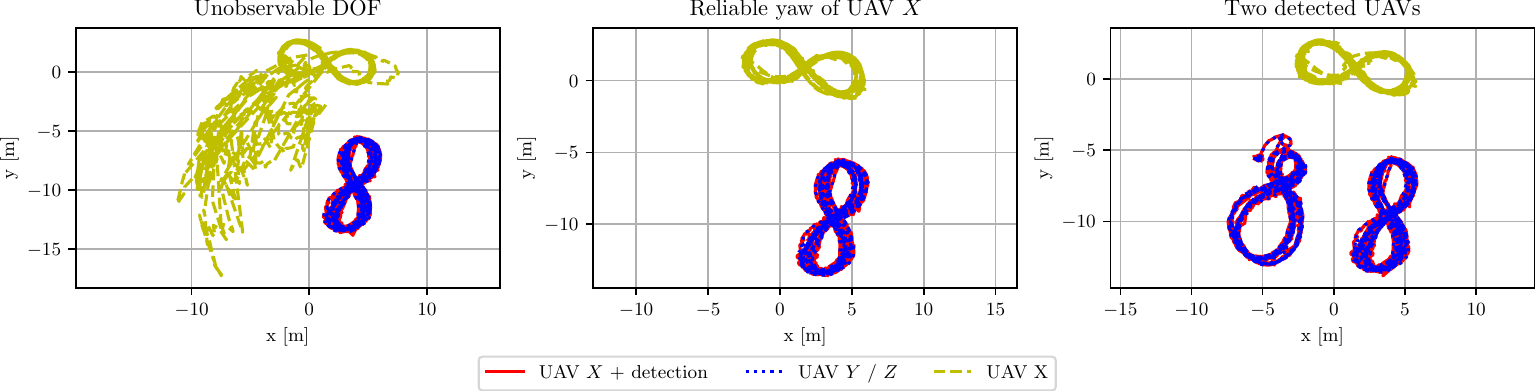}
  \caption{The effect of the unobservable direction during \ac{LIO} degradation. The left plot shows the situation with one detected \ac{UAV} and the unconstrained unobservable direction. The central plot shows the same situation when the yaw of \ac{UAV} $X$ is always considered reliable.
  The right plot shows the same situation as the left plot, but with two detected \acp{UAV}.
  }
  \label{fig:exp_obs_puav}
\end{figure*}

\noindent As established theoretically in sec.~\ref{sec:observability}, in the case of odometry degradations, the tackled cooperative localization problem exhibits unobservable directions.
In case of \ac{LIO} degradation, i.e., odometry degradation of the detecting robot $X$, there is one unobservable \ac{DOF} in the direction of the yaw of robot $X$ and the vector perpendicular to the line connecting the robots.
In case of \ac{VIO} degradation, i.e., odometry degradation of the detected robot $Y$, the yaw orientation of robot $Y$ is unobservable.
We explore these findings experimentally to verify them and to provide further insight into the problem.

\subsubsection{Unobservable DOF of the Detecting Robot $X$}

We utilized data from the experiment \textit{Outdoor \#3}, where one detecting \ac{UAV} $X$ and two detected \acp{UAV} $Y$ and $Z$ were following figure-eight trajectories.
We overwrote the eigenvalues of the \ac{LIO}'s approximate Hessian to consider the \ac{LIO} output degenerated between $t=\SI{300}{s}$ and $t=\SI{400}{s}$.
We compared three different situations.
First, we utilized only a single detected \ac{UAV} $Y$ in the cooperative localization process.
Second, we utilized only \ac{UAV} $Y$ and changed $\prescript{\mathrm{hi}}{}{\sigma}_\gamma$ to the same value as $\prescript{\mathrm{lo}}{}{\sigma}_\gamma$, forcing the algorithm to consider the yaw values of the \ac{LIO} measurements as reliable for the entire time.
Third, we utilized both of the detected \acp{UAV} $Y$ and $Z$.

Fig.~\ref{fig:exp_obs_puav} shows data from the analysis.
To analyze the consistency of information in the factor graph, we composed each \ac{UAV} detection with the pose of \ac{UAV} $X$ at the corresponding timestamp, and compared it with the estimated poses of the detected \acp{UAV} $Y$ and $Z$, interpolated to the same timestamp.
Furthermore, we plotted the estimated positions of \ac{UAV} $X$ at the same timestamp.
The left plot in Fig.~\ref{fig:exp_obs_puav} shows that there is no significant discrepancy between the detections composed with the poses of \ac{UAV} $X$ and the interpolated poses of \ac{UAV} $Y$, although the poses of \ac{UAV} $X$ exhibit significant drift.
The position drift is caused by the unobservable direction, as the estimated poses of \ac{UAV} $X$ fulfill the optimization objective to minimize the error of the detection factors and the error of the \ac{VIO} measurement factors, but at the time of the \ac{LIO} degeneracy, the poses of the \ac{UAV} $X$ are not fully constrained, resulting in the observed drift.

In the central plot of Fig.~\ref{fig:exp_obs_puav}, we constrained the yaw measurements from the \ac{LIO} algorithm to always be considered reliable.
Such a constraint eliminated the unobservable direction, and the position drift was mitigated.
In the right plot of Fig.~\ref{fig:exp_obs_puav}, we instead utilized two detected \acp{UAV} at the same time.
The position drift of \ac{UAV} $X$ disappeared, because the unobservable direction was mitigated by utilizing multiple detected \acp{UAV}, confirming the findings of the observability analysis from sec.~\ref{sec:obs_lio}.

Note that in the \textit{Outdoor \#1} experiment, which used a single detected \ac{UAV} and contained large areas of \ac{LIO} degradation, the drift caused by the unobservable direction was not as pronounced, as the \acp{UAV} were often changing their relative position, resulting in changes in the unobservability direction, and the errors caused by the unobservability were much smaller than the localization improvements provided by the cooperative localization method.

\subsubsection{Yaw Unobservability of the Detected Robot $Y$}
\begin{figure}[t]
  \centering
  \includegraphics[width=1.0\linewidth, trim=0.0cm 0.0cm 0.0cm 0.0cm, clip=true]{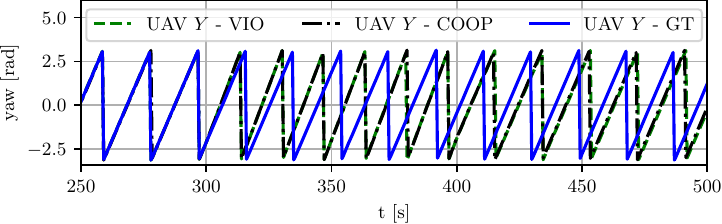}
  \caption{The effect of unobservable yaw of the detected \ac{UAV} $Y$ in the presence of pure yaw drift.
  We artificially inserted constant-velocity yaw drift between $t=\SI{300}{s}$ and $t=\SI{400}{s}$.
  }
  \label{fig:exp_obs_suav_yaw}
\end{figure}

We utilized data from the \textit{Indoor \#1} experiment.
Between $t=\SI{300}{s}$ and $t=\SI{400}{s}$, we artificially inserted cumulative drift of the yaw orientation of \ac{UAV} $Y$ with a constant velocity of \SI{0.05}{rad\per s}.
Fig.~\ref{fig:exp_obs_suav_yaw} shows a comparison of the resulting yaw of \ac{VIO}, the yaw estimated by the cooperative localization method, and the ground-truth yaw.
The yaw from \ac{VIO} and from the cooperative localization method coincide the entire time, and there is an increasing offset from the ground-truth data.
The offset stays constant even after the drift stops being added at $t=\SI{400}{s}$ because the yaw orientation of \ac{UAV} $Y$ is unobservable from other data than the \ac{VIO} measurements.
This behavior corresponds to the findings of the theoretical analysis in sec.~\ref{sec:obs_vio}.

\subsection{Discussion of Results}

\noindent The experiments have confirmed the ability of the proposed cooperative localization method to adapt to changing conditions and significantly improve the localization accuracy in the presence of sensory degradations.
The quantitative evaluation of experiments in sec.~\ref{sec:exp_turku} and sec.~\ref{sec:exp_temesvar} has shown the localization accuracy of the cooperative localization method to be similar to the accuracy of the individual robot odometries when no sensory degradations are present and to significantly improve the localization performance in the presence of sensory degradations.
Simultaneously, the method provides localization in a common reference frame, which is crucial for enabling any cooperation between the robots.
As shown in sec.~\ref{sec:exp_wasserstein}, the Wasserstein distance between the covariance matrices of consecutive \ac{VIO} positions highly correlates with the real-world localization error, especially in the presence of visual degradations.
Weighting the relative \ac{VIO} measurements proportionally to the Wasserstein distance is thus crucial for the adaptivity of the algorithm.
The benefits of utilizing the Wasserstein distance were clearly demonstrated in the ablation study, comparing it with the use of static variance values.
Finally, the findings of the theoretical analysis regarding the unobservable directions in the presence of odometry degradations were experimentally verified in sec.~\ref{sec:exp_observability}.
The experiments have confirmed that there is one unobservable \ac{DOF} in the presence of odometry degradation of the detecting robot, which can be mitigated by utilizing multiple detected robots.
Finally, the experiments have demonstrated the unobservability of the yaw orientation of the detected \ac{UAV}.
To mitigate this unobservability, the method would need to fuse an additional source of information about the yaw of the detected robot, or the detected robot itself would need to have the ability to detect other robots.



\section{Conclusions}
\label{sec:conclusion}

\noindent A novel multi-modal multi-robot adaptive cooperative localization approach was proposed in this paper.
The proposed method fuses \ac{LIO} and \ac{VIO} measurements from different robots with direct 3D detections between the robots in a loosely-coupled fashion using a factor graph-based formulation.
A novel interpolation-based quaternary factor enables efficient fusion of the data from unsynchronized sources of measurements.
The approach adaptively reacts to the changing reliability of the measurements to enhance the localization performance in the presence of sensory degradations.
The degradation of \ac{LIO} measurements is detected by analyzing the eigenvalues of the approximate Hessian of the scan matching problem.
The changing uncertainty of \ac{VIO} measurements is evaluated in a novel approach based on the Wasserstein distance between the covariance matrices of consecutive \ac{VIO} positions.
Theoretical analysis was performed to analyze the observability of the tackled cooperative localization problem under various conditions, especially in the presence of odometry degradations.
The proposed method and the findings of the theoretical analysis were extensively evaluated on real-world data gathered with a \ac{UGV}-\ac{UAV} and a \ac{UAV}-only team.
The accuracy of the approach was quantitatively evaluated with respect to motion capture and \ac{RTK} ground truth.
The experiments have shown that the proposed cooperative localization method provides significant improvements in localization accuracy in the presence of sensory degradations.


\printcredits


\begin{acronym}
  \acro{GPS}[GPS]{Global Positioning System}
  \acro{CNN}[CNN]{Convolutional Neural Network}
  \acro{MAV}[MAV]{Micro Aerial Vehicle}
  \acro{UAV}[UAV]{Unmanned Aerial Vehicle}
  \acro{UGV}[UGV]{Unmanned Ground Vehicle}
  \acro{UV}[UV]{ultraviolet}
  \acro{UVDAR}[\emph{UVDAR}]{UltraViolet Direction And Ranging}
  \acro{UT}[UT]{Unscented Transform}
  \acro{GNSS}[GNSS]{Global Navigation Satellite System}
  \acro{RTK}[RTK]{Real-Time Kinematic}
  \acro{MOCAP}[mo-cap]{Motion capture}
  \acro{ROS}[ROS]{Robot Operating System}
  \acro{MPC}[MPC]{Model Predictive Control}
  \acro{MBZIRC}[MBZIRC 2020]{Mohamed Bin Zayed International Robotics Challenge 2020}
  \acro{MBZIRC19}[MBZIRC 2019]{Mohamed Bin Zayed International Robotics Challenge 2019}
  \acro{FOV}[FOV]{Field Of View}
  \acrodefplural{FOV}[FOVs]{Fields of View}
  \acro{ICP}[ICP]{Iterative closest point}
  \acro{FSM}[FSM]{Finite-State Machine}
  \acro{IMU}[IMU]{Inertial Measurement Unit}
  \acro{EKF}[EKF]{Extended Kalman Filter}
  \acro{LKF}[LKF]{Linear Kalman Filter}
  \acro{POMDP}[POMDP]{Partially Observable Markov Decision Process}
  \acro{KF}[KF]{Kalman Filter}
  \acro{COTS}[COTS]{Commercially Available Off-the-Shelf}
  \acro{ESC}[ESC]{Electronic Speed Controller}
  \acro{lidar}[LiDAR]{Light Detection and Ranging}
  \acro{SLAM}[SLAM]{Simultaneous Localization and Mapping}
  \acro{SEF}[SEF]{Successive Edge Following}
  \acro{IEPF}[IEPF]{Iterative End-Point Fit}
  \acro{USAR}[USAR]{Urban Search and Rescue}
  \acro{SAR}[SAR]{Search and Rescue}
  \acro{ROI}[ROI]{Region of Interest}
  \acro{WEC}[WEC]{Window Edge Candidate}
  \acro{UAS}[UAS]{Unmanned Aerial System}
  \acro{VIO}[VIO]{Visual-Inertial Odometry}
  \acro{DOF}[DOF]{Degree of Freedom}
  \acrodefplural{DOF}[DOFs]{Degrees of Freedom}
  \acro{LTI}[LTI]{Linear Time-Invariant}
  \acro{FCU}[FCU]{Flight Control Unit}
  \acro{UWB}[UWB]{Ultra-wideband}
  \acro{ICP}[ICP]{Iterative Closest Point}
  \acro{NIS}[NIS]{Normalized Innovations Squared}
  \acro{LRF}[LRF]{Laser Rangefinder}
  \acro{RMSE}[RMSE]{Root Mean Squared Error}
  \acro{VINS}[VINS]{Vision-aided Inertial Navigation Systems}
  \acro{VSLAM}[VSLAM]{Visual Simultaneous Localization and Mapping}
  \acro{NLS}[NLS]{Non-linear Least Squares}
  \acro{NTP}[NTP]{Network Time Protocol}
  \acro{ATE}[ATE]{Absolute Trajectory Error}
  \acro{PUAV}[pUAV]{primary UAV}
  \acro{SUAV}[sUAV]{secondary UAV}
  \acro{NTP}[NTP]{Network Time Protocol}
  \acro{LOS}[LOS]{line-of-sight}
  \acro{MAE}[MAE]{Mean Absolute Error}
  \acro{LIO}[LIO]{LiDAR-Inertial Odometry}
  \acro{NEES}[NEES]{Normalized Estimation Error Squared}
  \acro{RPE}[RPE]{Relative Pose Error}
\end{acronym}


\bibliographystyle{cas-model2-names}

\bibliography{main}

@article{ebadiPresentFutureSLAM2024,
	title = {Present and {Future} of {SLAM} in {Extreme} {Environments}: {The} {DARPA} {SubT} {Challenge}},
	volume = {40},
	issn = {1941-0468},
	shorttitle = {Present and {Future} of {SLAM} in {Extreme} {Environments}},
	doi = {10.1109/TRO.2023.3323938},
	journal = {IEEE Transactions on Robotics},
	author = {Ebadi, Kamak and Bernreiter, Lukas and Biggie, Harel and Catt, Gavin and Chang, Yun and Chatterjee, Arghya and Denniston, Christopher E. and Deschênes, Simon-Pierre and Harlow, Kyle and Khattak, Shehryar and Nogueira, Lucas and Palieri, Matteo and Petráček, Pavel and Petrlík, Matěj and Reinke, Andrzej and Krátký, Vít and Zhao, Shibo and Agha-mohammadi, Ali-akbar and Alexis, Kostas and Heckman, Christoffer and Khosoussi, Kasra and Kottege, Navinda and Morrell, Benjamin and Hutter, Marco and Pauling, Fred and Pomerleau, François and Saska, Martin and Scherer, Sebastian and Siegwart, Roland and Williams, Jason L. and Carlone, Luca},
	year = {2024},
	pages = {936--959},
}

@inproceedings{khattakComplementaryMultiModal2020,
	title = {Complementary {Multi}–{Modal} {Sensor} {Fusion} for {Resilient} {Robot} {Pose} {Estimation} in {Subterranean} {Environments}},
	doi = {10.1109/ICUAS48674.2020.9213865},
	booktitle = {2020 {International} {Conference} on {Unmanned} {Aircraft} {Systems} ({ICUAS})},
	author = {Khattak, Shehryar and Nguyen, Huan and Mascarich, Frank and Dang, Tung and Alexis, Kostas},
	month = sep,
	year = {2020},
	pages = {1024--1029},
}

@inproceedings{zhaoSuperOdometryIMUcentric2021a,
	title = {Super {Odometry}: {IMU}-centric {LiDAR}-{Visual}-{Inertial} {Estimator} for {Challenging} {Environments}},
	shorttitle = {Super {Odometry}},
	doi = {10.1109/IROS51168.2021.9635862},
	booktitle = {2021 {IEEE}/{RSJ} {International} {Conference} on {Intelligent} {Robots} and {Systems} ({IROS})},
	author = {Zhao, Shibo and Zhang, Hengrui and Wang, Peng and Nogueira, Lucas and Scherer, Sebastian},
	month = sep,
	year = {2021},
	pages = {8729--8736},
}

@article{zhangLaserVisualInertial2018b,
	title = {Laser–visual–inertial odometry and mapping with high robustness and low drift},
	volume = {35},
	issn = {1556-4967},
	doi = {10.1002/rob.21809},
	language = {en},
	number = {8},
	journal = {Journal of Field Robotics},
	author = {Zhang, Ji and Singh, Sanjiv},
	year = {2018},
	pages = {1242--1264},
}

@article{zhuSwarmLIO2DecentralizedEfficient2025,
	title = {Swarm-{LIO2}: {Decentralized} {Efficient} {LiDAR}-{Inertial} {Odometry} for {Aerial} {Swarm} {Systems}},
	volume = {41},
	issn = {1941-0468},
	shorttitle = {Swarm-{LIO2}},
	doi = {10.1109/TRO.2024.3522155},
	journal = {IEEE Transactions on Robotics},
	author = {Zhu, Fangcheng and Ren, Yunfan and Yin, Longji and Kong, Fanze and Liu, Qingbo and Xue, Ruize and Liu, Wenyi and Cai, Yixi and Lu, Guozheng and Li, Haotian and Zhang, Fu},
	year = {2025},
	pages = {960--981},
}

@article{horynaFastSwarmingUAVs2024,
	title = {Fast {Swarming} of {UAVs} in {GNSS}-{Denied} {Feature}-{Poor} {Environments} {Without} {Explicit} {Communication}},
	volume = {9},
	issn = {2377-3766},
	doi = {10.1109/LRA.2024.3390596},
	number = {6},
	journal = {IEEE Robotics and Automation Letters},
	author = {Horyna, Jiří and Krátký, Vít and Pritzl, Václav and Báča, Tomáš and Ferrante, Eliseo and Saska, Martin},
	month = jun,
	year = {2024},
	pages = {5284--5291},
}

@article{pritzlFusionVisualInertialOdometry2023,
  author={Pritzl, Václav and Vrba, Matouš and Štěpán, Petr and Saska, Martin},
  journal={IEEE Access}, 
  title={Fusion of Visual-Inertial Odometry With LiDAR Relative Localization for Cooperative Guidance of a Micro-Scale Aerial Vehicle}, 
  year={2026},
  volume={14},
  number={},
  pages={31269-31285},
  doi={10.1109/ACCESS.2026.3666998}
}

@inproceedings{zhongCoLRIOLiDARRangingInertialCentralized2024,
	title = {{CoLRIO}: {LiDAR}-{Ranging}-{Inertial} {Centralized} {State} {Estimation} for {Robotic} {Swarms}},
	shorttitle = {{CoLRIO}},
	doi = {10.1109/ICRA57147.2024.10611672},
	booktitle = {2024 {IEEE} {International} {Conference} on {Robotics} and {Automation} ({ICRA})},
	author = {Zhong, Shipeng and Chen, Hongbo and Qi, Yuhua and Feng, Dapeng and Chen, Zhiqiang and Wu, Jin and Wen, Weisong and Liu, Ming},
	month = may,
	year = {2024},
	pages = {3920--3926},
}

@article{xuOmniSwarmDecentralizedOmnidirectional2022,
	title = {Omni-{Swarm}: {A} {Decentralized} {Omnidirectional} {Visual}–{Inertial}–{UWB} {State} {Estimation} {System} for {Aerial} {Swarms}},
	issn = {1941-0468},
	shorttitle = {Omni-{Swarm}},
	doi = {10.1109/TRO.2022.3182503},
	journal = {IEEE Transactions on Robotics},
	author = {Xu, Hao and Zhang, Yichen and Zhou, Boyu and Wang, Luqi and Yao, Xinjie and Meng, Guotao and Shen, Shaojie},
	year = {2022},
  volume={38},
  number={6},
  pages={3374-3394},
}

@inproceedings{schmuckCOVINSVisualInertialSLAM2021,
	title = {{COVINS}: {Visual}-{Inertial} {SLAM} for {Centralized} {Collaboration}},
	shorttitle = {{COVINS}},
	doi = {10.1109/ISMAR-Adjunct54149.2021.00043},
	booktitle = {2021 {IEEE} {International} {Symposium} on {Mixed} and {Augmented} {Reality} {Adjunct} ({ISMAR}-{Adjunct})},
	author = {Schmuck, Patrik and Ziegler, Thomas and Karrer, Marco and Perraudin, Jonathan and Chli, Margarita},
	month = oct,
	year = {2021},
	pages = {171--176},
}

@article{zhongDCLSLAMDistributedCollaborative2024,
	title = {{DCL}-{SLAM}: {A} {Distributed} {Collaborative} {LiDAR} {SLAM} {Framework} for a {Robotic} {Swarm}},
	volume = {24},
	issn = {1558-1748},
	shorttitle = {{DCL}-{SLAM}},
	doi = {10.1109/JSEN.2023.3345541},
	number = {4},
	journal = {IEEE Sensors Journal},
	author = {Zhong, Shipeng and Qi, Yuhua and Chen, Zhiqiang and Wu, Jin and Chen, Hongbo and Liu, Ming},
	month = feb,
	year = {2024},
	pages = {4786--4797},
}

@article{huangDiSCoSLAMDistributedScan2022a,
	title = {{DiSCo}-{SLAM}: {Distributed} {Scan} {Context}-{Enabled} {Multi}-{Robot} {LiDAR} {SLAM} {With} {Two}-{Stage} {Global}-{Local} {Graph} {Optimization}},
	volume = {7},
	issn = {2377-3766},
	shorttitle = {{DiSCo}-{SLAM}},
	doi = {10.1109/LRA.2021.3138156},
	number = {2},
	journal = {IEEE Robotics and Automation Letters},
	author = {Huang, Yewei and Shan, Tixiao and Chen, Fanfei and Englot, Brendan},
	month = apr,
	year = {2022},
	pages = {1150--1157},
}

@article{heGroundAerialCollaborative2021,
	title = {Ground and {Aerial} {Collaborative} {Mapping} in {Urban} {Environments}},
	volume = {6},
	issn = {2377-3766},
	doi = {10.1109/LRA.2020.3032054},
	number = {1},
	journal = {IEEE Robotics and Automation Letters},
	author = {He, Jinhao and Zhou, Yuming and Huang, Lixiang and Kong, Yang and Cheng, Hui},
	month = jan,
	year = {2021},
	pages = {95--102},
}

@article{changLAMP20Robust2022,
	title = {{LAMP} 2.0: {A} {Robust} {Multi}-{Robot} {SLAM} {System} for {Operation} in {Challenging} {Large}-{Scale} {Underground} {Environments}},
	volume = {7},
	issn = {2377-3766},
	shorttitle = {{LAMP} 2.0},
	doi = {10.1109/LRA.2022.3191204},
	number = {4},
	journal = {IEEE Robotics and Automation Letters},
	author = {Chang, Yun and Ebadi, Kamak and Denniston, Christopher E. and Ginting, Muhammad Fadhil and Rosinol, Antoni and Reinke, Andrzej and Palieri, Matteo and Shi, Jingnan and Chatterjee, Arghya and Morrell, Benjamin and Agha-mohammadi, Ali-akbar and Carlone, Luca},
	month = oct,
	year = {2022},
	pages = {9175--9182},
}

@article{lajoieSwarmSLAMSparseDecentralized2024,
	title = {Swarm-{SLAM}: {Sparse} {Decentralized} {Collaborative} {Simultaneous} {Localization} and {Mapping} {Framework} for {Multi}-{Robot} {Systems}},
	volume = {9},
	issn = {2377-3766},
	shorttitle = {Swarm-{SLAM}},
	doi = {10.1109/LRA.2023.3333742},
	number = {1},
	journal = {IEEE Robotics and Automation Letters},
	author = {Lajoie, Pierre-Yves and Beltrame, Giovanni},
	month = jan,
	year = {2024},
	pages = {475--482},
}

@article{tianKimeraMultiRobustDistributed2022a,
	title = {Kimera-{Multi}: {Robust}, {Distributed}, {Dense} {Metric}-{Semantic} {SLAM} for {Multi}-{Robot} {Systems}},
	volume = {38},
	issn = {1941-0468},
	shorttitle = {Kimera-{Multi}},
	doi = {10.1109/TRO.2021.3137751},
	number = {4},
	journal = {IEEE Transactions on Robotics},
	author = {Tian, Yulun and Chang, Yun and Herrera Arias, Fernando and Nieto-Granda, Carlos and How, Jonathan P. and Carlone, Luca},
	month = aug,
	year = {2022},
	pages = {2022--2038},
}

@INPROCEEDINGS{birdDVMSLAMDecentralizedVisual2025,
  author={Bird, Joshua and Blumenkamp, Jan and Prorok, Amanda},
  booktitle={2025 IEEE International Conference on Robotics and Automation (ICRA)}, 
  title={{DVM-SLAM: Decentralized Visual Monocular Simultaneous Localization and Mapping for Multi-Agent Systems}}, 
  year={2025},
  volume={},
  number={},
  pages={1-7},
  doi={10.1109/ICRA55743.2025.11127510}}

@article{tunaInformedConstrainedAligned2024a,
	title = {Informed, {Constrained}, {Aligned}: {A} {Field} {Analysis} on {Degeneracy}-aware {Point} {Cloud} {Registration} in the {Wild}},
	issn = {2997-1101},
	shorttitle = {Informed, {Constrained}, {Aligned}},
	doi = {10.1109/TFR.2025.3576053},
	journal = {IEEE Transactions on Field Robotics},
	author = {Tuna, Turcan and Nubert, Julian and Pfreundschuh, Patrick and Cadena, Cesar and Khattak, Shehryar and Hutter, Marco},
	year = {2025},
  volume={2},
  pages={485-515},
}

@article{tunaXICPLocalizabilityAwareLiDAR2024b,
	title = {X-{ICP}: {Localizability}-{Aware} {LiDAR} {Registration} for {Robust} {Localization} in {Extreme} {Environments}},
	volume = {40},
	issn = {1941-0468},
	shorttitle = {X-{ICP}},
	doi = {10.1109/TRO.2023.3335691},
	journal = {IEEE Transactions on Robotics},
	author = {Tuna, Turcan and Nubert, Julian and Nava, Yoshua and Khattak, Shehryar and Hutter, Marco},
	year = {2024},
	pages = {452--471},
}

@inproceedings{shanLIOSAMTightlycoupledLidar2020,
	title = {{LIO}-{SAM}: {Tightly}-coupled {Lidar} {Inertial} {Odometry} via {Smoothing} and {Mapping}},
	shorttitle = {{LIO}-{SAM}},
	doi = {10.1109/IROS45743.2020.9341176},
	booktitle = {2020 {IEEE}/{RSJ} {International} {Conference} on {Intelligent} {Robots} and {Systems} ({IROS})},
	author = {Shan, Tixiao and Englot, Brendan and Meyers, Drew and Wang, Wei and Ratti, Carlo and Rus, Daniela},
	month = oct,
	year = {2020},
	pages = {5135--5142},
}

@inproceedings{genevaOpenVINSResearchPlatform2020a,
	title = {{OpenVINS}: {A} {Research} {Platform} for {Visual}-{Inertial} {Estimation}},
	shorttitle = {{OpenVINS}},
	doi = {10.1109/ICRA40945.2020.9196524},
	booktitle = {2020 {IEEE} {International} {Conference} on {Robotics} and {Automation} ({ICRA})},
	author = {Geneva, Patrick and Eckenhoff, Kevin and Lee, Woosik and Yang, Yulin and Huang, Guoquan},
	month = may,
	year = {2020},
	pages = {4666--4672},
}

@article{zhengFASTLIVO2FastDirect2025,
	title = {{FAST}-{LIVO2}: {Fast}, {Direct} {LiDAR}–{Inertial}–{Visual} {Odometry}},
	volume = {41},
	issn = {1941-0468},
	shorttitle = {{FAST}-{LIVO2}},
	doi = {10.1109/TRO.2024.3502198},
	journal = {IEEE Transactions on Robotics},
	author = {Zheng, Chunran and Xu, Wei and Zou, Zuhao and Hua, Tong and Yuan, Chongjian and He, Dongjiao and Zhou, Bingyang and Liu, Zheng and Lin, Jiarong and Zhu, Fangcheng and Ren, Yunfan and Wang, Rong and Meng, Fanle and Zhang, Fu},
	year = {2025},
	pages = {326--346},
}

@article{yuanSRLIVOLiDARInertialVisualOdometry2024,
	title = {{SR}-{LIVO}: {LiDAR}-{Inertial}-{Visual} {Odometry} and {Mapping} {With} {Sweep} {Reconstruction}},
	volume = {9},
	issn = {2377-3766},
	shorttitle = {{SR}-{LIVO}},
	doi = {10.1109/LRA.2024.3389415},
	number = {6},
	journal = {IEEE Robotics and Automation Letters},
	author = {Yuan, Zikang and Deng, Jie and Ming, Ruiye and Lang, Fengtian and Yang, Xin},
	month = jun,
	year = {2024},
	pages = {5110--5117},
}

@article{linR33LIVERobustRealTime2024,
	title = {R$^{\textrm{3}}${3LIVE}++: {A} {Robust}, {Real}-{Time}, {Radiance} {Reconstruction} {Package} {With} a {Tightly}-{Coupled} {LiDAR}-{Inertial}-{Visual} {State} {Estimator}},
	volume = {46},
	issn = {1939-3539},
	shorttitle = {R$^{\textrm{3}}${3LIVE}++},
	doi = {10.1109/TPAMI.2024.3456473},
	number = {12},
	journal = {IEEE Transactions on Pattern Analysis and Machine Intelligence},
	author = {Lin, Jiarong and Zhang, Fu},
	month = dec,
	year = {2024},
	pages = {11168--11185},
}

@article{zhangLVIOFusionTightlyCoupledLiDARVisualInertial2024,
	title = {{LVIO}-{Fusion}:{Tightly}-{Coupled} {LiDAR}-{Visual}-{Inertial} {Odometry} and {Mapping} in {Degenerate} {Environments}},
	volume = {9},
	issn = {2377-3766},
	shorttitle = {{LVIO}-{Fusion}},
	doi = {10.1109/LRA.2024.3371383},
	number = {4},
	journal = {IEEE Robotics and Automation Letters},
	author = {Zhang, Hongkai and Du, Liang and Bao, Sheng and Yuan, Jianjun and Ma, Shugen},
	month = apr,
	year = {2024},
	pages = {3783--3790},
}

@inproceedings{graeterLIMOLidarMonocularVisual2018,
	title = {{LIMO}: {Lidar}-{Monocular} {Visual} {Odometry}},
	shorttitle = {{LIMO}},
	doi = {10.1109/IROS.2018.8594394},
	booktitle = {2018 {IEEE}/{RSJ} {International} {Conference} on {Intelligent} {Robots} and {Systems} ({IROS})},
	author = {Graeter, Johannes and Wilczynski, Alexander and Lauer, Martin},
	month = oct,
	year = {2018},
	pages = {7872--7879},
}

@article{xuIntermittentVIOAssistedLiDAR2025,
	title = {Intermittent {VIO}-{Assisted} {LiDAR} {SLAM} {Against} {Degeneracy}: {Recognition} and {Mitigation}},
	volume = {74},
	issn = {1557-9662},
	shorttitle = {Intermittent {VIO}-{Assisted} {LiDAR} {SLAM} {Against} {Degeneracy}},
	doi = {10.1109/TIM.2024.3507053},
	journal = {IEEE Transactions on Instrumentation and Measurement},
	author = {Xu, Jiahao and Li, Tuan and Wang, Hongxia and Wang, Zhipeng and Bai, Tong and Hou, Xiaopeng},
	year = {2025},
	pages = {1--13},
}

@inproceedings{shanLVISAMTightlycoupledLidarVisualInertial2021a,
	title = {{LVI}-{SAM}: {Tightly}-coupled {Lidar}-{Visual}-{Inertial} {Odometry} via {Smoothing} and {Mapping}},
	shorttitle = {{LVI}-{SAM}},
	doi = {10.1109/ICRA48506.2021.9561996},
	booktitle = {2021 {IEEE} {International} {Conference} on {Robotics} and {Automation} ({ICRA})},
	author = {Shan, Tixiao and Englot, Brendan and Ratti, Carlo and Rus, Daniela},
	month = may,
	year = {2021},
	pages = {5692--5698},
}

@inproceedings{zhangDegeneracyOptimizationbasedState2016a,
	title = {On degeneracy of optimization-based state estimation problems},
	doi = {10.1109/ICRA.2016.7487211},
	booktitle = {2016 {IEEE} {International} {Conference} on {Robotics} and {Automation} ({ICRA})},
	author = {Zhang, Ji and Kaess, Michael and Singh, Sanjiv},
	month = may,
	year = {2016},
	pages = {809--816},
}

@article{wangMCLIVOLowdriftLiDARinertialvisual2025a,
	title = {{MCLIVO}: {A} low-drift {LiDAR}-inertial-visual odometry with multi-constrained optimization for planetary mapping},
	volume = {240},
	issn = {0263-2241},
	shorttitle = {{MCLIVO}},
	doi = {10.1016/j.measurement.2024.115551},
	journal = {Measurement},
	author = {Wang, Yankun and Yao, Weiran and Zhang, Bing and Sun, Guanghui and Zheng, Bo and Cao, Tao},
	month = jan,
	year = {2025},
	pages = {115551},
}

@article{leeMINSEfficientRobust2023,
	title = {{MINS}: {Efficient} and {Robust} {Multisensor}-{Aided} {Inertial} {Navigation} {System}},
	issn = {1556-4967},
	shorttitle = {{MINS}},
	doi = {10.1002/rob.22546},
	language = {en},
	journal = {Journal of Field Robotics},
	author = {Lee, Woosik and Geneva, Patrick and Chen, Chuchu and Huang, Guoquan},
	year = {2025},
}

@article{mangelsonCharacterizingUncertaintyJointly2020,
	title = {Characterizing the {Uncertainty} of {Jointly} {Distributed} {Poses} in the {Lie} {Algebra}},
	volume = {36},
	issn = {1941-0468},
	doi = {10.1109/TRO.2020.2994457},
	number = {5},
	journal = {IEEE Transactions on Robotics},
	author = {Mangelson, Joshua G. and Ghaffari, Maani and Vasudevan, Ram and Eustice, Ryan M.},
	month = oct,
	year = {2020},
	pages = {1371--1388},
}

@article{umeyamaLeastsquaresEstimationTransformation1991,
	title = {Least-squares estimation of transformation parameters between two point patterns},
	volume = {13},
	issn = {1939-3539},
	doi = {10.1109/34.88573},
	number = {4},
	journal = {IEEE Transactions on Pattern Analysis and Machine Intelligence},
	author = {Umeyama, S.},
	month = apr,
	year = {1991},
	pages = {376--380},
}

@article{solaMicroLieTheory2021,
	title = {A micro {Lie} theory for state estimation in robotics},
	journal = {arXiv:1812.01537 [cs]},
	author = {Solà, Joan and Deray, Jeremie and Atchuthan, Dinesh},
	year = {2021},
}

@article{vrbaOnboardLiDARBasedFlying2025,
	title = {On {Onboard} {LiDAR}-{Based} {Flying} {Object} {Detection}},
	volume = {41},
	issn = {1941-0468},
	doi = {10.1109/TRO.2024.3502494},
	journal = {IEEE Transactions on Robotics},
	author = {Vrba, Matouš and Walter, Viktor and Pritzl, Václav and Pliska, Michal and Báča, Tomáš and Spurný, Vojtěch and Heřt, Daniel and Saska, Martin},
	year = {2025},
	pages = {593--611},
}

@article{hertMRSDroneModular2023a,
	title = {{MRS} {Drone}: {A} {Modular} {Platform} for {Real}-{World} {Deployment} of {Aerial} {Multi}-{Robot} {Systems}},
	volume = {108},
	issn = {1573-0409},
	shorttitle = {{MRS} {Drone}},
	doi = {10.1007/s10846-023-01879-2},
	language = {en},
	number = {4},
	journal = {Journal of Intelligent \& Robotic Systems},
	author = {Hert, Daniel and Baca, Tomas and Petracek, Pavel and Kratky, Vit and Penicka, Robert and Spurny, Vojtech and Petrlik, Matej and Vrba, Matous and Zaitlik, David and Stoudek, Pavel and Walter, Viktor and Stepan, Petr and Horyna, Jiri and Pritzl, Vaclav and Sramek, Martin and Ahmad, Afzal and Silano, Giuseppe and Licea, Daniel Bonilla and Stibinger, Petr and Nascimento, Tiago and Saska, Martin},
	month = jul,
	year = {2023},
	pages = {64},
}

@article{dellaertFactorGraphsRobot2017,
	title = {Factor {Graphs} for {Robot} {Perception}},
	volume = {6},
	issn = {1935-8253, 1935-8261},
	doi = {10.1561/2300000043},
	language = {en},
	number = {1-2},
	journal = {Foundations and Trends in Robotics},
	author = {Dellaert, Frank and Kaess, Michael},
	year = {2017},
	pages = {1--139},
}

@article{barfootAssociatingUncertaintyThreeDimensional2014,
	title = {Associating {Uncertainty} {With} {Three}-{Dimensional} {Poses} for {Use} in {Estimation} {Problems}},
	volume = {30},
	issn = {1941-0468},
	doi = {10.1109/TRO.2014.2298059},
	number = {3},
	journal = {IEEE Transactions on Robotics},
	author = {Barfoot, Timothy D. and Furgale, Paul T.},
	month = jun,
	year = {2014},
	pages = {679--693},
}

@article{kaessISAM2IncrementalSmoothing2012,
	title = {{iSAM2}: {Incremental} smoothing and mapping using the {Bayes} tree},
	volume = {31},
	issn = {0278-3649},
	shorttitle = {{iSAM2}},
	doi = {10.1177/0278364911430419},
	language = {en},
	number = {2},
	journal = {The International Journal of Robotics Research},
	author = {Kaess, Michael and Johannsson, Hordur and Roberts, Richard and Ila, Viorela and Leonard, John J and Dellaert, Frank},
	month = feb,
	year = {2012},
	pages = {216--235},
}

@article{olkinDistanceTwoRandom1982,
	title = {The distance between two random vectors with given dispersion matrices},
	volume = {48},
	issn = {0024-3795},
	doi = {10.1016/0024-3795(82)90112-4},
	journal = {Linear Algebra and its Applications},
	author = {Olkin, I. and Pukelsheim, F.},
	month = dec,
	year = {1982},
	pages = {257--263},
}

@article{dowsonFrechetDistanceMultivariate1982a,
	title = {The {Fréchet} distance between multivariate normal distributions},
	volume = {12},
	issn = {0047-259X},
	doi = {10.1016/0047-259X(82)90077-X},
	number = {3},
	journal = {Journal of Multivariate Analysis},
	author = {Dowson, D. C and Landau, B. V},
	month = sep,
	year = {1982},
	pages = {450--455},
}

@article{bhatiaBuresWassersteinDistance2019,
	title = {On the {Bures}–{Wasserstein} distance between positive definite matrices},
	volume = {37},
	issn = {0723-0869},
	doi = {10.1016/j.exmath.2018.01.002},
	number = {2},
	journal = {Expositiones Mathematicae},
	author = {Bhatia, Rajendra and Jain, Tanvi and Lim, Yongdo},
	month = jun,
	year = {2019},
	pages = {165--191},
}

@inproceedings{zhangTutorialQuantitativeTrajectory2018,
	title = {A {Tutorial} on {Quantitative} {Trajectory} {Evaluation} for {Visual}(-{Inertial}) {Odometry}},
	doi = {10.1109/IROS.2018.8593941},
	booktitle = {2018 {IEEE}/{RSJ} {International} {Conference} on {Intelligent} {Robots} and {Systems} ({IROS})},
	author = {Zhang, Zichao and Scaramuzza, Davide},
	month = oct,
	year = {2018},
}

@INPROCEEDINGS{Spasojevic-RSS-23, 
    AUTHOR    = {Igor Spasojevic AND Xu Liu AND Alejandro Ribeiro AND George J. Pappas AND Vijay Kumar}, 
    TITLE     = {{Active Collaborative Localization in Heterogeneous Robot Teams}}, 
    BOOKTITLE = {Proceedings of Robotics: Science and Systems}, 
    YEAR      = {2023}, 
    MONTH     = {July}, 
    DOI       = {10.15607/RSS.2023.XIX.112} 
}

\bio{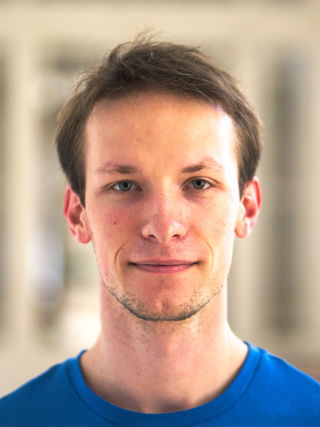}
V\'aclav Pritzl received the Ph.D. degree in Informatics from the Czech Technical University in Prague (CTU in Prague), Czech Republic, in 2026, on the topic of Cooperative UAV Navigation in GNSS-Denied Environments.
He is a member of the Multi-robot Systems Group, CTU in Prague.
  He has authored or coauthored 13 publications in conferences and impacted journals with >300 citations indexed by Scholar and h-index 10. His research interests include cooperative navigation of teams of UAVs in GNSS-denied environments. He was a member of CTU-UPENN-NYU team in the MBZIRC 2020.
\endbio

\bio{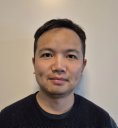}
Xianjia Yu received the D.Sc. (Tech.) degree in Robotics and Autonomous Systems from the University of Turku, Finland, in 2024, with a focus on multi-modal sensor fusion and perception. He is currently a Postdoctoral Researcher with the Turku Intelligent Embedded and Robotic Systems (TIERS) group at the University of Turku. His research interests include multi-modal sensing and perception, machine learning in robotics, and multi-robot systems. He is also an Automation Researcher and Engineer at Kaptas Oy, Turku, Finland.
\endbio

\bio{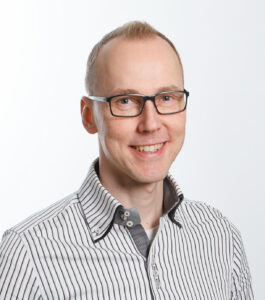}
Tomi Westerlund is a professor in robotics and autonomous systems. He is the research group leader of the Turku Intelligent Embedded and Robotics Systems (TIERS) lab at the University of Turku (UTU), Finland. His main research interests are in collaborative and heterogeneous multi-robot systems and autonomous robots in urban and unstructured environments. In essence, to better understand how robots operate and can perform a variety of functions in different fields like environmental monitoring, delivery, and search and rescue missions. Westerlund has >200 peer-reviewed publications. He is a senior member of IEEE.
\endbio

\bio{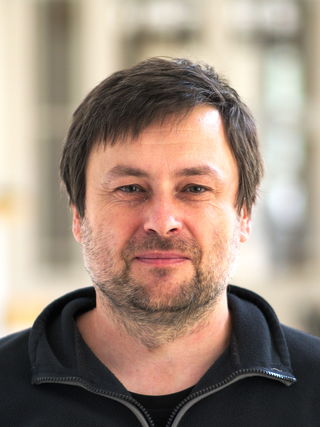}
Petr \v{S}t\v{e}p\'an received the Ph.D. degree in sensor fusion for mapping from the Czech Technical University in Prague (CTU Prague), Prague, Czech Republic, in 2002.
He is currently with the Multi-Robot Systems lab, CTU Prague, where he focuses on sensor fusion, mapping, localization, and planning for unmanned aerial vehicles.
He has also been involved in industrial projects and the H2020 AerialCore project.
He is a coauthor of >40 publications in conferences and impacted journals with >800 citations indexed by Scholar and h-index 12. He was a member of CTU-UPenn-UoL and CTU-UPENN-NYU teams in the MBZIRC 2017 and MBZIRC 2020 robotic competitions in Abu Dhabi.
\endbio

\bio{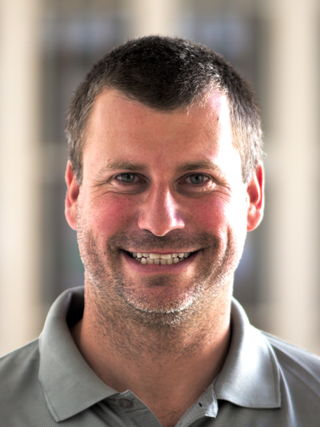}
Martin Saska received the Ph.D. degree in trajectory planning and optimal control for formations of autonomous robots from the University of Wuerzburg, Wuerzburg, Germany, in 2010, within the Ph.D. program of Elite Network of Bavaria.
He founded and heads the Multi-robot Systems group at the Czech Technical University in Prague with more than 40 researchers.
He was a Visiting Scholar with the University of Illinois at Urbana-Champaign, Champaign, IL, USA, and with the University of Pennsylvania, Philadelphia, PA, USA.
He has authored or coauthored >200 publications in conferences and impacted journals, including IJRR, AURO, JFR, ASC, EJC, with >9000 citations indexed by Scholar and h-index 53.
His team won multiple robotic challenges in MBZIRC 2017, MBZIRC 2020, and DARPA SubT competitions.
\endbio


\end{document}